\documentclass[lettersize,journal]{IEEEtran}
\usepackage{times}
\usepackage{epsfig}
\usepackage{graphicx}
\usepackage{amsmath}
\usepackage{amssymb}
\usepackage{multirow}
\usepackage{makecell}
\usepackage{url} 
\usepackage{color}
\usepackage[most]{tcolorbox}
\usepackage{array}
\usepackage{tabularx}
\usepackage[T1]{fontenc}
\usepackage[utf8]{inputenc}
\usepackage{textcomp}
\usepackage{url}
\usepackage{graphicx}
\usepackage{multirow}
\usepackage{booktabs}
\usepackage{colortbl}
\usepackage{booktabs}
\usepackage{amssymb}
\usepackage{graphicx}      
\usepackage[table,xcdraw]{xcolor} 
\usepackage{pifont}
\usepackage{graphicx}
\newcommand{\xmark}{\includegraphics[height=1.7ex]{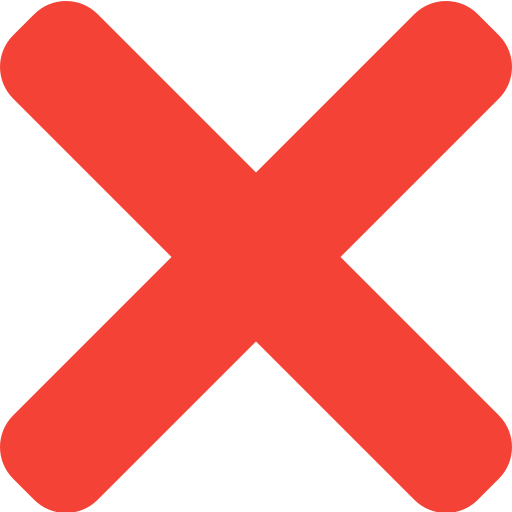}}
\newcommand{\cmark}{\includegraphics[height=2ex]{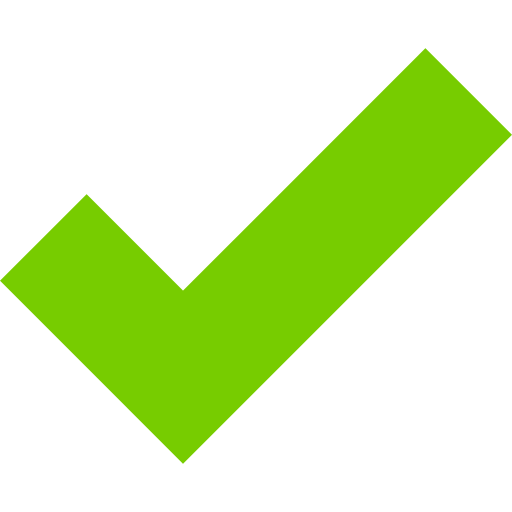}}

\usepackage{tcolorbox}
\usepackage{microtype}
\usepackage{amssymb}
\usepackage{pifont}
\usepackage{hyperref} 
\definecolor{darkblue}{RGB}{208, 219, 221} 
\definecolor{lightgreen}{RGB}{237, 242, 240} 
\definecolor{myblue}{RGB}{150, 150, 230}
\definecolor{darkgray}{RGB}{169,169,169}   
\definecolor{myblue}{RGB}{0,102,204}
\definecolor{lightgray}{RGB}{230,230,230}  %

\begin{document}

\title{Beyond Empathy: Integrating Diagnostic and Therapeutic Reasoning with Large Language Models for Mental Health
Counseling}

\author{
He Hu,~Juzheng Si,~Qianning Wang,~Fei Ma,~Fuji Ren,~\IEEEmembership{Senior Member},~Sicheng Zhao,~\IEEEmembership{Senior Member},\\Yucheng Zhou,~and~Laizhong Cui,~\IEEEmembership{Senior Member}%
\thanks{He Hu is with the College of Computer Science and Software Engineering, Shenzhen University, Shenzhen, China (email: huhe@gml.ac.cn).}%
\thanks{Juzheng Si is with Shandong University. Email: sijuzheng@mail.sdu.edu.cn}%
\thanks{Qianning Wang is with Department of Computer Science, Faculty of Science and Engineering, University of Nottingham Ningbo China. Email: qianning.wang@nottingham.edu.cn}%
\thanks{Fei Ma is with the Guangdong Laboratory of Artificial Intelligence and Digital Economy (SZ). Emails: mafei@gml.ac.cn}%
\thanks{Fuji Ren is with the School of Computer Science and Engineering, University of Electronic Science and Technology of China, Chengdu, 611731, China. (email:
renfuji@uestc.edu.cn).}%
\thanks{Sicheng Zhao is with the Department of Psychological and Cognitive Sciences, Tsinghua University, Beijing, 100084, China (email: schzhao@tsinghua.edu.cn).}%
\thanks{Yucheng Zhou is with SKL-IOTSC, Department of Artificial Intelligence, University of Macau. Email: yucheng.zhou@connect.um.edu.mo.}%
\thanks{Laizhong Cui is with the College of Computer Science and Software Engineering, Shenzhen University, Shenzhen, China (email: cuilz@szu.edu.cn).}%
\thanks{Yucheng Zhou and Laizhong Cui are the corresponding authors. Emails: yucheng.zhou@connect.um.edu.mo, cuilz@szu.edu.cn.}%
}

%


\markboth{}
{Shell \MakeLowercase{\textit{et al.}}: Bare Demo of IEEEtran.cls for Computer Society Journals}

\IEEEcompsoctitleabstractindextext{%
\begin{abstract}
Large language models (LLMs) hold significant potential for psychological support, capable of generating empathetic responses and simulating therapeutic conversations. However, existing LLM-based approaches often lack sufficient grounding in structured diagnostic reasoning and diverse therapeutic strategies, limiting their effectiveness in complex counseling scenarios. To address these limitations, we propose PsyLLM, the first large language model designed to incorporate diagnostic- and therapeutic-oriented reasoning for mental health counseling. To develop PsyLLM, we design a novel automated data synthesis pipeline that processes real-world mental health posts collected from Reddit, where users frequently express psychological distress and seek community support. The pipeline transforms these posts into multi-turn counseling dialogues and leverages LLMs guided by international diagnostic standards (e.g., DSM/ICD) and multiple therapeutic frameworks (e.g., CBT, ACT, psychodynamic) to generate structured, clinically informed reasoning traces. A rigorous multi-dimensional filtering mechanism further improves the quality and clinical fidelity of the synthesized dialogue data. In addition, we introduce a new benchmark and evaluation protocol, assessing counseling performance across four key dimensions. Our experiments, along with evaluations by professional counseling experts, demonstrate that PsyLLM consistently outperforms various baseline models on this benchmark. Code, data, and models are available at \url{https://github.com/Emo-gml/PsyLLM}.
\end{abstract}

\begin{IEEEkeywords}
Mental Health, Large Language Models, Language Agents, Computational Psychiatry, Longitudinal Support
\end{IEEEkeywords}}

\maketitle
\IEEEdisplaynotcompsoctitleabstractindextext
\IEEEpeerreviewmaketitle
\section{Introduction}
\label{sec:introduction}

\begin{figure}[!t]
    \centering
    \includegraphics[width=1\linewidth]{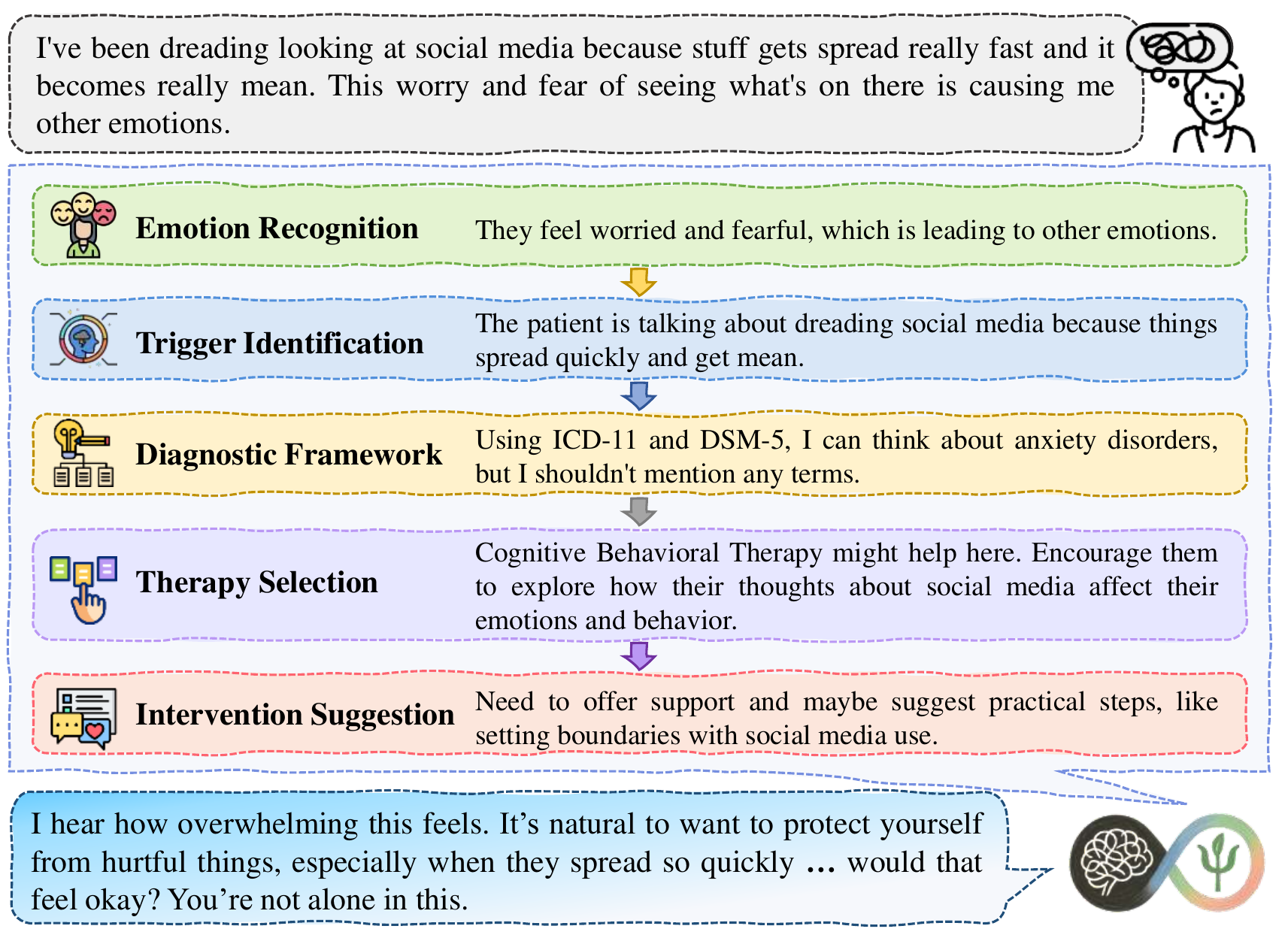}
    \vspace{-7mm}
    \caption{\small PsyLLM simulates therapeutic reasoning by assessing emotions, analyzing cognitive patterns, and formulating strategies grounded in DSM/ICD criteria and diverse modalities (e.g., CBT, ACT, psychodynamic). This enables clinically informed, context-sensitive counseling responses.}
    \label{fig:intro}
\end{figure}

\IEEEPARstart{L}{arge} language models (LLMs) have demonstrated promising capabilities in psychological support tasks, including recognizing emotional distress, generating empathetic responses, and facilitating therapeutic conversations~\cite{ identifying,annaagent,zhang2026tutorial}. Recent studies have further explored affect-aware conversational agents that incorporate emotional appraisal, perceived empathy, and cognitive reappraisal to provide more interpretable and context-sensitive support~\cite{bhattacharjee2026user,zhang2026mentalcare}. However, existing systems often rely on surface-level emotional cues and lack structured reasoning over users' psychological states, underlying concerns, and appropriate therapeutic strategies, limiting their effectiveness in complex psychological counseling scenarios~\cite{Psychotherapy,raamkumar2022empathetic,minddialog}.

\begin{table*}[t]\small
\centering
\caption{\small Comparison of OpenR1-Psy with existing psychological dialogue datasets, including whether each dataset is publicly available.}
\setlength{\tabcolsep}{1.7mm}
\begin{tabular}{lcccccc}
\toprule
\textbf{Dataset} &
\textbf{Public} &
\textbf{Turn Type} &
\textbf{Reasoning?} &
\textbf{Therapy-Guided} &
\textbf{Clinically-Guided} &
\textbf{Dialogues} \\
\midrule
Psych8k \cite{Chatcounselor1}   & \cmark & Single-turn         & \xmark & \xmark & \xmark & 8,187 \\
CPsyCounD \cite{CPsyCoun}      & \cmark & Multi-turn          & \xmark & \cmark & \xmark & 3,134 \\
SweetieChat \cite{SweetieChat} & \xmark & Multi-turn          & \xmark & \xmark & \xmark & 3,757 \\
PsyDTCorpus \cite{PsyDT1} & \xmark  & Multi-turn          & \xmark & \xmark & \xmark & 5,000 \\
TeenEmpathSupport \cite{TeenEmpath}             & \xmark & Multi-turn          & \xmark & \xmark & \xmark &  6,590 \\
MDD-5k \cite{MD5k}             & \xmark & Multi-turn          & \xmark & \xmark & \cmark & 5,000 \\
\rowcolor{blue!15}
\textbf{OpenR1-Psy (Ours)}     & \cmark & Single + Multi-turn & \cmark & \cmark & \cmark & 19,302 \\
\bottomrule
\end{tabular}
\label{tab:dataset}
\end{table*}

A clinically grounded counseling system must account for both diagnostic rigor and therapeutic diversity. However, current LLM-based approaches fall short in the following two critical aspects:
\textbf{(1) Lack of deep reasoning on explicit diagnosis.} In clinical practice, mental health professionals rely on international diagnostic standards such as the \textit{Diagnostic and Statistical Manual of Mental Disorders (DSM)} and the \textit{International Classification of Diseases (ICD)} to assess symptom patterns, determine diagnoses, and guide treatment planning \cite{DSM-5,icd}. These standards ensure consistency, explainability, and legitimacy, especially in complex or comorbid cases. However, current counseling LLMs do not reason explicitly according to such frameworks, limiting their reliability in clinical decision support or realistic therapy simulation.
\textbf{(2) Neglect of therapeutic diversity.} Not all patients benefit from the same therapeutic approach \cite{lingxidiagbench,castonguay2015research,beutler2006systematic}. Depending on the individual's context, symptoms, and preferences, therapists may adopt different modalities such as cognitive behavioral therapy (CBT), acceptance and commitment therapy (ACT), psychodynamic therapy, or humanistic approaches \cite{rowan2006humanistic, CBT,wang2026care}. This diversity is essential for effective care, but current models rarely incorporate multiple therapeutic perspectives or adapt their strategies accordingly.

To address existing limitations, we propose the first large language model that systematically integrates both diagnostic and therapeutic reasoning for mental health counseling, as shown in Fig.~\ref{fig:intro}. Specifically, the model is capable of autonomously applying diagnostic criteria and various psychotherapeutic frameworks during counseling interactions. As shown in Table~\ref{tab:dataset}, we introduce a novel data synthesis pipeline that: (1) automatically parses real-world mental health posts; (2) generates simplified multi-turn diagnostic dialogues; (3) guides LLMs to simulate the reasoning processes of mental health professionals, incorporating step-by-step thinking aligned with DSM/ICD diagnostic logic and therapeutic considerations drawn from multiple evidence-based treatments; and (4) applies multi-dimensional filtering, including Incomplete Thinking, Context Incoherence, Mismatch Response, and No Framework, to ensure data quality. The resulting high-quality filtered dataset is used to fine-tune the LLM, enabling it to deliver counseling responses that are not only accurate and therapeutically grounded but also contextually appropriate and aligned with clinical reasoning processes.

In our experiments, we construct a benchmark to evaluate the quality of counseling responses along four key dimensions: Empathy \& Insight, Support \& Autonomy, Attunement \& Presence, Safety \& Boundaries. Furthermore, we analyze the training data used for counseling responses, revealing that while increased model scale improves performance, the importance of data quantity is highly dependent on its quality.
Our main contributions are as follows:
\begin{itemize}
    \item We propose PsyLLM, the first LLM that systematically integrates both diagnostic and therapeutic reasoning for mental health counseling.
    \item We develop an automated data synthesis pipeline that generates high-quality, multi-turn dialogues reflecting diagnostic and therapeutic reasoning.
    \item We incorporate international diagnostic standards (e.g., DSM/ICD) and diverse therapeutic strategies (e.g., CBT, ACT, psychodynamic) to guide the model's deep reasoning.
    \item We introduce a new benchmark and evaluation protocol tailored for counseling quality, and conduct a comprehensive analysis of model performance and data characteristics.
\end{itemize}

\section{Background and Application}\label{sec:background}
\subsection{Psychological Counseling Chatbots}
The growing demand for accessible and scalable mental health support has fueled the development of AI-driven dialogue systems capable of offering emotional and psychological assistance \cite{ge2025survey,belli2026vera,pattern}. Early efforts in this area primarily centered on single-turn, empathy-oriented response generation. For instance, ChatCounselor \cite{Chatcounselor1} trained LLMs to produce emotionally appropriate responses evaluated across human-centric metrics. Similarly, PsyQA \cite{PsyQA} focused on generating supportive long-form answers in Chinese, but remained restricted to single-turn interactions without modeling therapeutic intent or multi-turn dynamics.
To move beyond isolated responses, later work explored multi-turn emotional support. SMILE \cite{SMILE} proposed a framework that expands single supportive utterances into coherent, multi-turn conversations using ChatGPT, enabling better contextual continuity. SoulChat \cite{SoulChat} extended this by fine-tuning LLMs on multi-turn empathetic dialogues, improving their ability to adapt across evolving emotional scenarios.
Advancing further, CPsyCoun \cite{CPsyCoun} reconstructed multi-turn dialogues with therapeutic type annotations (e.g., CBT, humanistic), though it lacked modeling of the underlying reasoning behind those therapeutic strategies. 

Recent work has broadened AI-based mental health support beyond dialogue generation. Some studies incorporate IoT-based emotion feedback into psychological interventions and use smartphone-derived behavioral features for depression detection \cite{echoes, digital,depguard}. Other studies focus on safety-aligned mental health risk intervention \cite{secure}. Therabot has been evaluated in a randomized trial of a generative AI chatbot for mental health treatment \cite{randomized}. A cognitive-layer architecture augments general-purpose LLMs with specialized psychotherapeutic reasoning and has been evaluated in CBT interactions \cite{cognitive}. In contrast, PsyLLM focuses on constructing OpenR1-Psy and fine-tuning a model to jointly generate diagnostic and therapeutic reasoning with counseling responses in text-based, multi-turn interactions.
\begin{figure*}[t]
    \centering
    \includegraphics[width=1\linewidth]{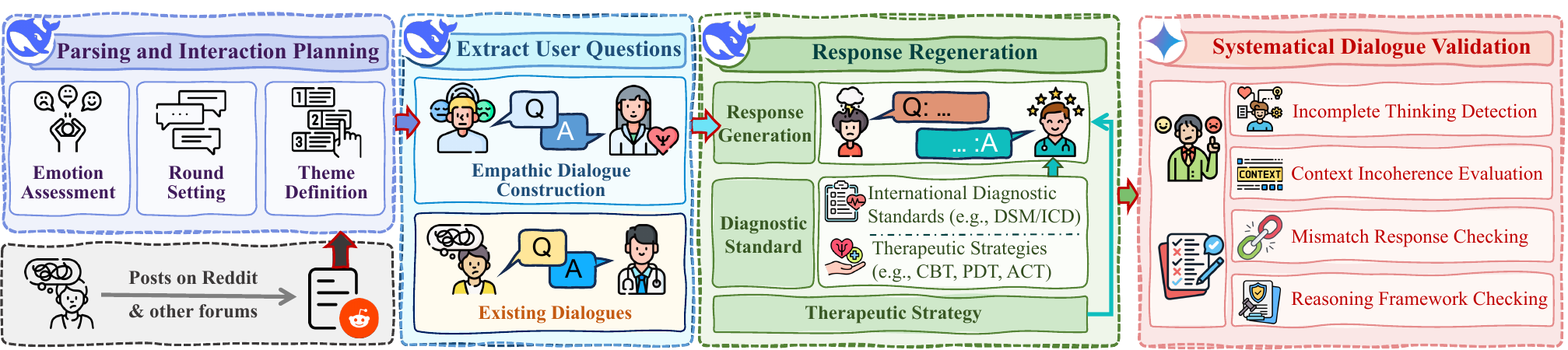}
    \caption{\small Overview of the OpenR1-Psy dataset construction pipeline. The process includes five stages: (1) Data collection from Reddit and real counseling datasets. (2) Parsing and interaction planning using a language model to assess emotions, define dialogue rounds, and set therapeutic themes. (3) Extraction of patient utterances from simulated dialogues and real counseling data. (4) Generation of reasoning traces and counselor responses based on diagnostic standards and therapeutic strategies. (5) Multi-dimensional validation to ensure coherence, clinical relevance, and reasoning quality.}
    \label{fig:pipeline}
\end{figure*}

\subsection{Deep Reasoning in LLMs}
Deep reasoning in LLMs has attracted increasing attention, particularly since the introduction of CoT prompting \cite{kardia,theramind,wang2026}, which encourages models to generate intermediate reasoning steps before producing final answers. This technique has shown notable improvements on arithmetic and symbolic tasks, revealing the latent ability of LLMs to perform multi-step inference. Recent work has extended deep reasoning capabilities beyond symbolic computation into more complex domains such as math reasoning, medical understanding, multimodal reasoning, and emotional understanding \cite{wang2026explainable,hu2025emobench,zhou2024visual}. For example,  Zhou et al.~\cite{dou2026dsadf} propose a dual-system framework inspired by Kahneman’s theory of fast and slow thinking, integrating reinforcement learning and vision-language reasoning to support adaptive decision-making. Similarly, VidEmo~\cite{VidEmo} leverages affective-tree reinforcement learning to enhance video emotion reasoning, progressively integrating attribute perception, expression analysis, and high-level emotional understanding capabilities. In mathematical contexts, Wang et al.~\cite{wang2025mathcoder} introduce a model that bridges vision and code by aligning image understanding with code generation, achieving superior performance in structured geometry problems compared to models like GPT-4o.
As research pushes toward more interactive and human-centric applications, deep reasoning in dialogue systems has emerged as a challenging yet crucial frontier. Maintaining coherence, adaptability, and emotional sensitivity across multiple conversational turns introduces new demands. To address this,  Xie et al.~\cite{xie2025leveraging} propose an LPE framework that combines CoT with emotional perception from spoken input, thereby guiding emotionally intelligent response generation. Mind AI's Mind~\cite{wu2025mind} uses LLMs to generate clinically aligned explanations for depression diagnosis.

\section{OpenR1-Psy Dataset Construction}
The construction of the OpenR1-Psy dataset follows a systematic pipeline (Fig.~\ref{fig:pipeline}), integrating data collection with LLM-powered parsing, therapeutic reasoning, and rigorous validation. We will detail each key component of the dataset creation process.

\subsection{Data Collection} 
To ensure both clinical realism and contextual diversity, our dataset sources are primarily divided into two parts: 
(1) The Reddit posts we used were pre-filtered and sourced from three publicly available corpora in mental health-related communities: Identifying-depression \cite{Identifying1}, Dreaddit \cite{Dreaddit}, and LRF \cite{Annotated}.
(2) Authentic patient utterances are derived from two high-quality psychological counseling datasets, i.e., ChatCounselor \cite{Chatcounselor1} and CPsyCoun \cite{CPsyCoun}, which are based on real-world clinical cases and authentic patient-doctor dialogues, and refined as multi-turn dialogues between patients and LLM counselors by ChatGPT \cite{GPT4}.

Rather than relying on isolated sentences, the post data we used are long-form posts that provide rich background information, including the poster's personal context, psychological state, and social environment. These posts span diverse life situations, such as family dynamics, friendship challenges, interpersonal conflict, psychiatric symptoms, and financial stressors. In total, we collected 10,097 posts from these sources. After removing duplicates based on exact textual matching, we retained 8,752 unique posts. This curated dataset serves as the foundation for our subsequent experiments on psychological state inference and mental health modeling.

\subsection{Data Parsing and Interaction Planning}

To convert raw Reddit posts into structured inputs for guiding dialogue generation, we implement a parsing and planning process inspired by psychological counseling techniques. This process uses a large language model (LLM), guided by a framework that simulates a counselor's initial assessment and planning upon receiving a client's self-report.  Specifically, we use DeepSeek \cite{deepseek} for this parsing and planning process. The released pipeline uses deepseek-chat for post analysis, dialogue construction, and quality filtering, and deepseek-reasoner for counselor reasoning and response generation. The scripts specify a temperature of 1.0 for generation and 0 for filtering; the prompts and implementation are provided in the Appendix and project repository. For each Reddit post, the LLM performs the following analyses: 
(1) \textbf{Emotion Assessment:} Identify the user's primary emotions, their intensity, and any emotional nuances. This step parallels how a therapist assesses affect. The output captures the key emotional states expressed.
(2) \textbf{Round Setting:} Estimate the number of dialogue turns (1 to 3) based on the complexity of the issue. The dialogue progresses from surface-level emotions to deeper needs or vulnerabilities, reflecting typical therapeutic stages.
(3) \textbf{Theme Definition:} Define the therapeutic focus for each round. These themes guide the conversation’s goals, ensuring each round builds on the previous one.

\subsection{Question Extraction from Dialogue}
Following the initial parsing and interaction planning, we prepare inputs for the deep reasoning process by creating a collection of patient utterances that will serve as prompts for counselor response generation. Let $\mathcal{U}_P$ denote this collection of patient utterances. This is achieved through two mechanisms:
(1) For the processed Reddit posts, we simulate a simplified initial empathetic multi-turn exchange using an LLM. Based on the planned dialogue structure (incorporating emotion assessment $E$, round setting $N$, and theme definition $Th$), the LLM first generates a hypothetical empathetic counselor response to the post, followed by a plausible patient utterance in reaction to this turn. This patient's utterance, representing a question or continuation of the user's expression, is extracted as a prompt for the subsequent deep reasoning step. This effectively transforms static post data $P_{raw}$ into dynamic dialogue starters, contributing to $\mathcal{U}_P$.
(2) We incorporate data from existing high-quality psychological counseling datasets, specifically ChatCounselor \cite{Chatcounselor1} and CPsyCoun \cite{CPsyCoun}. These datasets $\mathcal{D}_{exist}$ contain multi-turn patient-counselor dialogues. We extract only the patient utterances from these datasets, which also contribute to $\mathcal{U}_P$.

The collection of patients' turns $\mathcal{U}_P$ from both sources forms the input pool for the subsequent generation step. This approach is central to our methodology: instead of utilizing pre-existing counselor responses, the collected patient utterances serve as prompts that guide the LLM to first generate a deep reasoning trace, which is then used to construct the final counselor response.

\subsection{Response Generation with Diagnostic Standard \& Therapeutic Strategy}
Given a patient utterance $U_P \in \mathcal{U}_P$ (and potentially preceding dialogue context $C$), our system generates a clinically grounded counselor response $U_C$. Unlike methods that directly map $U_P$ to $U_C$ and might offer post-hoc explanations, our approach explicitly models the internal reasoning process concurrently with generating the response. The LLM produces outputs that include both the reasoning trace $R$ and the counselor response $U_C$ in a single generation step, systematically guided by key principles of clinical practice: international diagnostic standards and diverse therapeutic strategies.

The integration of these elements is crucial for effective and responsible mental health support. International diagnostic standards, such as the \textit{Diagnostic and Statistical Manual of Mental Disorders (DSM)} \cite{DSM-5} and the \textit{International Classification of Diseases (ICD)} \cite{icd}, provide a structured framework for understanding symptom patterns, recognizing potential underlying conditions, and ensuring a degree of diagnostic rigor that is fundamental in clinical assessment. Diverse therapeutic strategies, including but not limited to Cognitive Behavioral Therapy (CBT) \cite{CBT}, Acceptance and Commitment Therapy (ACT) \cite{hayes2005acceptance}, and psychodynamic approaches \cite{rowan2006humanistic}, offer varied theoretical lenses and practical techniques to address the patient's issues in a personalized and theoretically informed manner. Relying on a single strategy can limit applicability, as different patients respond better to different modalities. The data-generation prompt (Figure~\ref{fig:Therapeutic}) guides the model to select therapeutic approaches according to the client's needs while maintaining empathy and emotional validation. An integrative response may draw on multiple approaches, as illustrated by the combination of CBT-oriented reframing and humanistic self-compassion in Section~\ref{sec:case-study}.

Our system leverages a deep reasoning LLM to simulate this multi-faceted clinical thought process. Given $U_P$ and context $C$, the LLM generates a structured output containing both the reasoning trace $R$ and the counselor utterance $U_C$ by considering relevant diagnostic perspectives (informed by $D$, a representation of diagnostic knowledge) and evaluating applicable therapeutic strategies (informed by $T$, a representation of therapeutic knowledge). This combined generation process can be conceptually represented as:
\begin{align} \label{eq:combined_generation}
    \!\!(R, U_C) \!= \!\text{LLM}_{\text{DeepReasoning}}(U_P, C; D, T)
\end{align}
Here, $D$ and $T$ act as guiding frameworks that the LLM utilizes during the generation process. The model is specifically trained to first articulate the simulated clinical assessment and therapeutic rationale in $R$, and then formulate the final counselor response $U_C$ conditioned on $U_P$, $C$, and the generated $R$ within this single output structure.

This joint generation of $R$ and $U_C$ ensures that the generated counseling response $U_C$ is not merely empathetic or plausible but is also demonstrably grounded in the clinical reasoning captured by $R$. The reasoning $R$ is explicitly informed by diagnostic standards $D$ and therapeutic strategies $T$, providing a level of interpretability and clinical legitimacy that is often lacking in end-to-end generative models. This aligns the LLM's behavior more closely with the structured thought process of a human mental health professional, where the therapeutic intervention (the response) is directly derived from and justified by the clinical assessment and theoretical formulation (the reasoning).
\begin{figure*}[!t]
    \centering
    \includegraphics[width=0.31\linewidth]{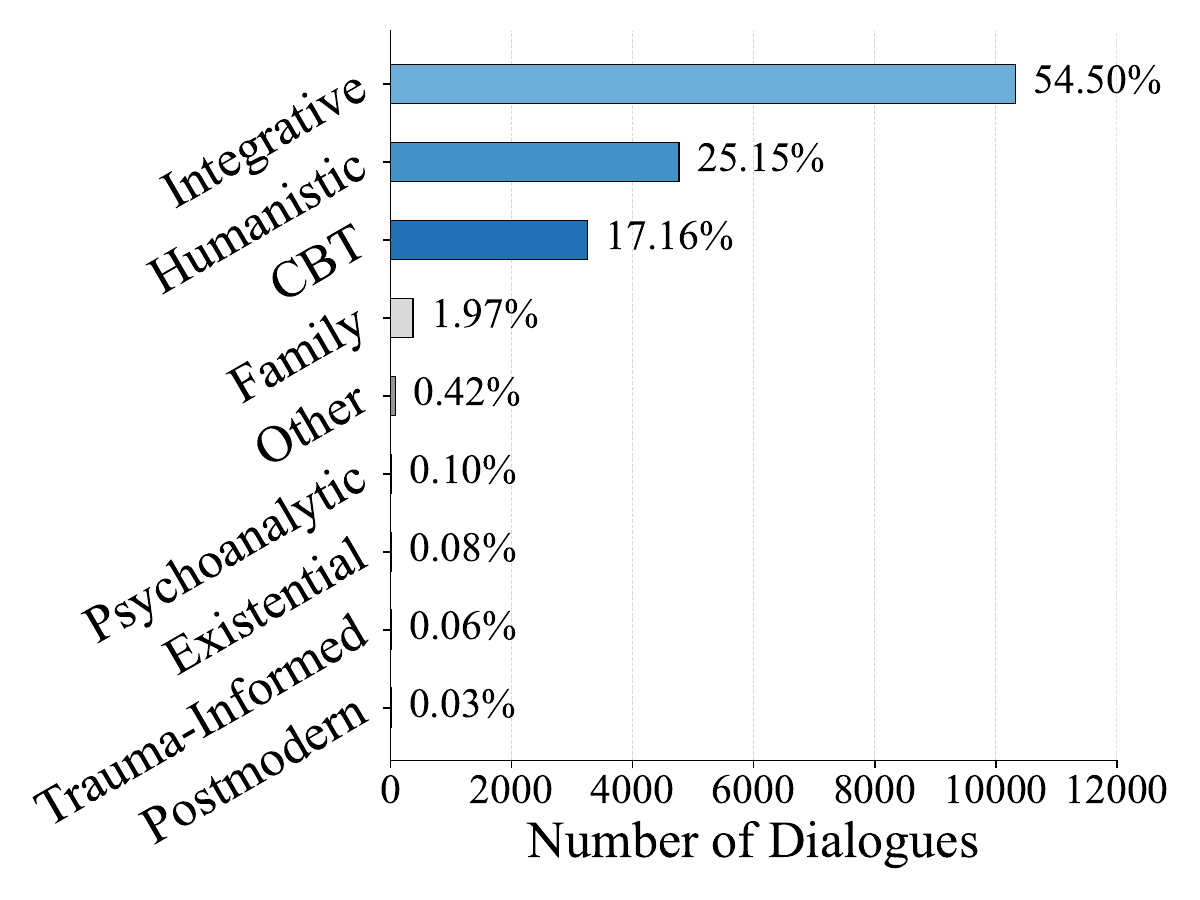} 
    \includegraphics[width=0.39\linewidth]{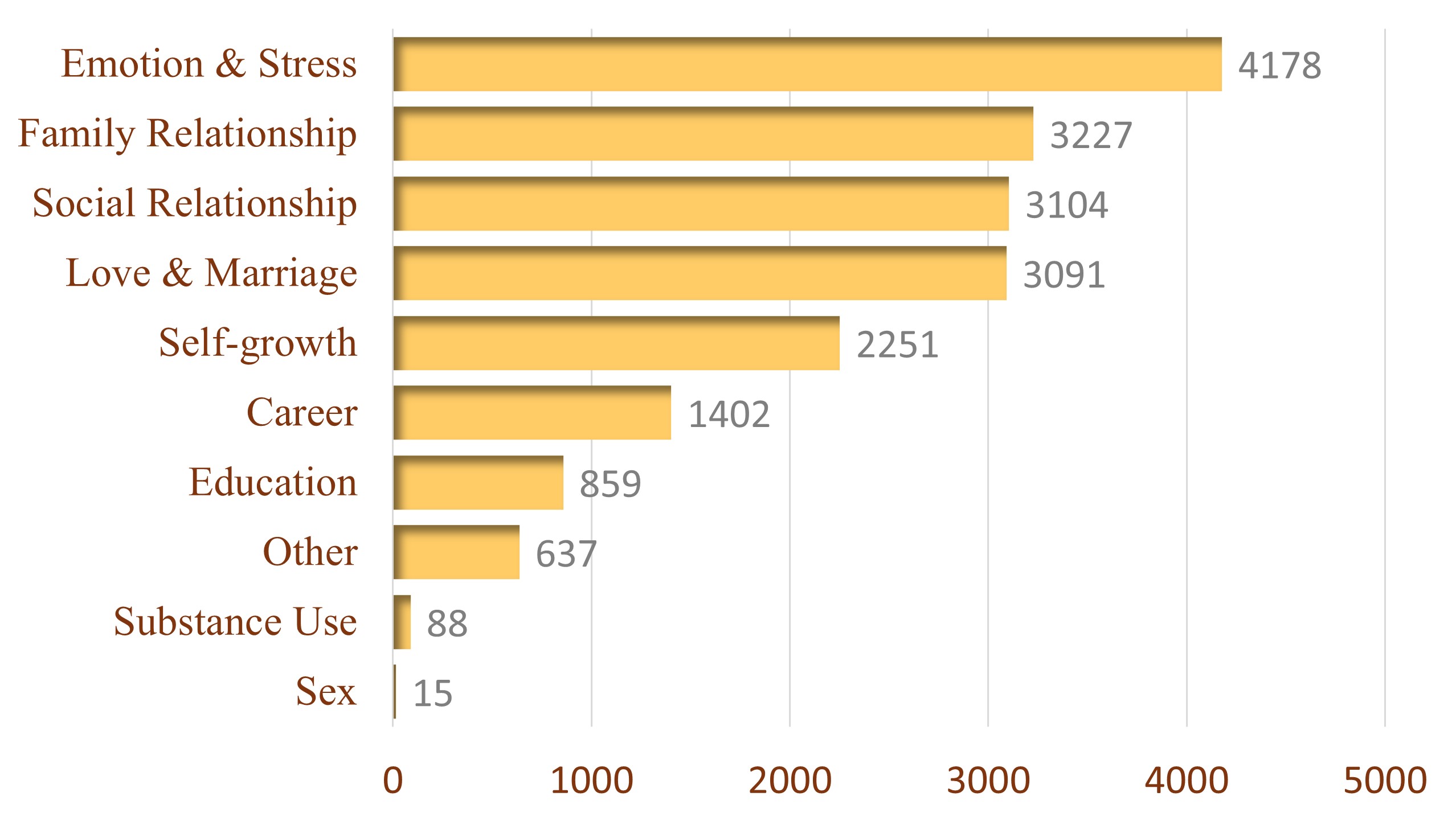}
    \includegraphics[width=0.285\linewidth]{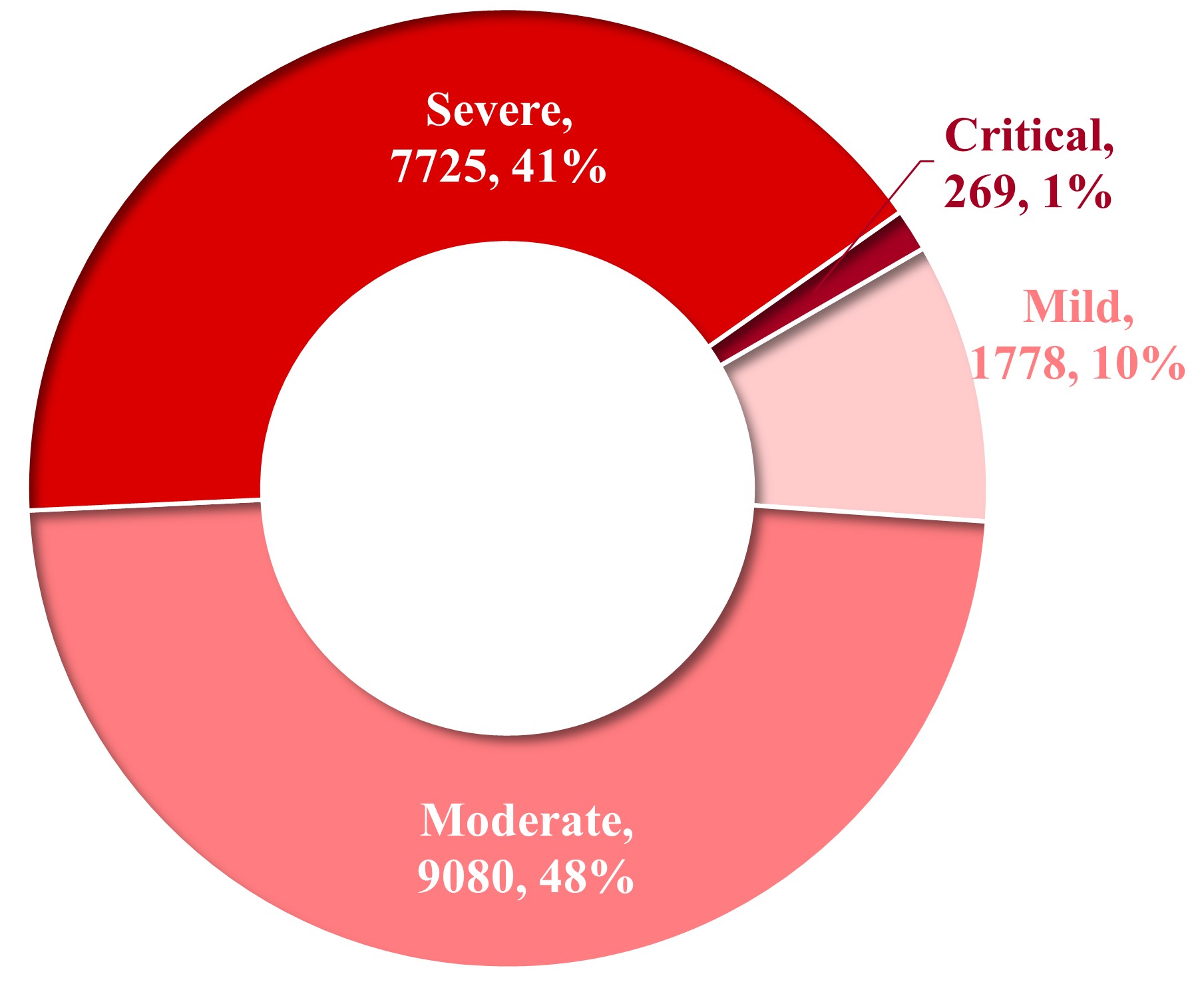}
    \caption{\small Analysis of OpenR1-Psy Dataset: (Left) Distribution of Psychotherapy Approaches; (Middle) Distribution of Scene Categories; (Right) Distribution of Severity Levels.}
    \label{fig:Distribution}
\end{figure*}

\subsection{Systematical Dialogue Validation}
\label{sec:dialogue-validation}
After the generation process, synthesized dialogue turns are constructed by combining patient utterances $U_P$ with their corresponding generated reasoning traces $R$ and counselor responses $U_C$. To ensure the clinical relevance, logical consistency, and overall quality of this data for fine-tuning PsyLLM, we implement a rigorous validation process. This validation employs a multi-dimensional filtering mechanism applied to each generated turn $(R, U_C)$ based on $U_P$ and context $C$. A sample is kept if it satisfies the boolean condition $K$:
\begin{align}
K = C_1 \land C_2 \land C_3 \land C_4
\end{align}
where the boolean variables $C_i$ represent the evaluation along four key dimensions: $C_1$, Complete Thinking Detection, which assesses the completeness and logical flow of the reasoning trace $R$; $C_2$, Context Coherence Evaluation, which evaluates the consistency of $(R, U_C)$ with the preceding dialogue context $C$ and the user prompt $U_P$; $C_3$, Mismatch Response Checking, which verifies the alignment between the reasoning trace $R$ and the counselor response $U_C$; and $C_4$, Reasoning Framework Checking, which confirms whether $R$ adheres to the intended guidance from international diagnostic standards ($D$) and therapeutic strategies ($T$).
Through this process, low-quality generated segments are filtered out, resulting in a high-quality dataset used for training. 
This systematic quality control is essential for building a reliable and clinically informed mental health counseling system.

\subsection{Dataset Statistics and Analysis}
\begin{table}[!t]\small
\centering
\caption{\small Statistics of the OpenR1-Psy Dataset (Train / Test).}
\resizebox{\linewidth}{!}{
\setlength{\tabcolsep}{1.2mm}
\begin{tabular}{lrrr}
\toprule
\textbf{Category} & \textbf{Train} & \textbf{Test}  & \textbf{All}\\
\midrule
\# Dialogues & 18,852 & 450 & 19,302 \\
\# Utterances & 47,723 & 1,651 & 49,374\\
Avg. Multi-turns & 3.76 & 5.04 & 3.80 \\
Avg. Patient Utterance Length & 224.18 & 171.84 & 222.43 \\
Avg. Counselor Utterance Length & 520.11 & 489.4 & 519.08\\
Avg. Reasoning Utterance Length & 1,592.33 & 1,627.36 & 1,593.5 \\
\bottomrule
\end{tabular}
}
\label{tab:dataset_stats}
\end{table}

\begin{table*}[!t]\small 
\centering
\caption{\small Synthetic data comparison across four metrics and normalized averages based on metric max scores: 2, 4, 3, 1. OpenR1-Psy achieves the highest performance and substantial relative improvements.}
\setlength{\tabcolsep}{1mm}
\begin{tabular}{lccccc}
\toprule
\multirow{2}{*}{\textbf{Method}} & \multicolumn{5}{c}{\textbf{Metrics}} \\
\cmidrule(lr){2-6}
& \textbf{Emp\&In} & \textbf{Sup\&Aut} & \textbf{Att\&Pre} & \textbf{Saf\&Bou} & \textbf{Normalized Avg} \\
\midrule
Psych8k \cite{Chatcounselor1}    & 0.740~~~~~~~~~~~~~~ & 1.830~~~~~~~~~~~~~ & 1.430~~~~~~~~~~~~~ & 0.990~~~~~~~~~~~ & 0.574~~~~~~~~~~~~~ \\
\rowcolor{blue!15}
\textbf{OpenR1-Psy} & \textbf{1.630} (+120\%)  & \textbf{3.590} (+96\%) & \textbf{2.660} (+86\%) & \textbf{1.000} (+1\%) & \textbf{0.900} (+57\%) \\
\midrule
CPsyCounD \cite{CPsyCoun}        & 0.950~~~~~~~~~~~~ & 1.950~~~~~~~~~~~~ & 1.840~~~~~~~~~~~~ & 0.980~~~~~~~~~~~ & 0.639~~~~~~~~~~~~ \\
\rowcolor{blue!15}
\textbf{OpenR1-Psy} & \textbf{1.630} (+71\%) & \textbf{3.590} (+84\%) & \textbf{2.660} (+44\%) & \textbf{1.000} (+2\%) & \textbf{0.900} (+41\%) \\
\bottomrule
\end{tabular}
\label{tab:data_result}
\end{table*}

We analyze the constructed OpenR1-Psy dataset, presenting its key statistics and characteristics in Table~\ref{tab:dataset_stats} and Fig.~\ref{fig:Distribution}.
Table~\ref{tab:dataset_stats} shows the dataset statistics, comprising 19,302 dialogues (49,374 utterances) with a train/test split. Patient utterances average 222 tokens, counselor responses 519 tokens, and notably, the reasoning trace averages 1593 tokens, highlighting its detailed nature.
Fig.~\ref{fig:Distribution} presents key dataset distributions. (Left) shows psychotherapy approaches, dominated by Integrative (54.50\%), Humanistic (25.15\%), and CBT (17.16\%), while including other diverse modalities. (Middle) illustrates scene categories, covering a broad range of topics like Emotion \& Stress, Family Relationship, and Social Relationship. (Right) depicts severity levels: Moderate (48\%) and Severe (41\%) are most frequent, alongside Mild (10\%) and Critical (1\%), representing varied intensity.

\subsection{PsyLLM Training}
This section details the training process for PsyLLM, our large language model specialized for psychological counseling, leveraging the high-quality OpenR1-Psy dataset. The primary objective is to train the model to generate clinically informed counselor responses ($U_C$) grounded in explicit therapeutic reasoning ($R$), given a patient's utterance ($U_P$) and preceding dialogue context ($C$).

We employ Supervised Fine-Tuning (SFT) on the OpenR1-Psy dataset. Each training instance uses the input sequence $(C, U_P)$ and the target output sequence $(R, U_C)$, where the validated reasoning trace $R$ is concatenated before the corresponding counselor response $U_C$. This structured format is crucial; it trains the model to first simulate the internal clinical reasoning process (generating $R$) and then formulate the response ($U_C$) conditioned on that reasoning and the input.

The training objective is to maximize the likelihood of generating the target sequence $(R, U_C)$ given $(C, U_P)$ by minimizing the cross-entropy loss over the target tokens. Formally, the model parameters $\theta$ are optimized by minimizing:
\begin{align} \label{eq:sft_loss_simplified}
\mathcal{L}(\theta)\!\! =\! -\frac{1}{N} \!\!\sum_{i=1}^{N}\! \sum_{t=1}^{|R_i U_{C,i}|} \!\!\log P_{\theta}(y_{i,t} | C_i, U_{P,i}, y_{i,<t})
\end{align}
where $N$ is the number of training instances, $|R_i U_{C,i}|$ is the length of the concatenated target sequence for instance $i$, $y_{i,t}$ is the $t$-th token, and $y_{i,<t}$ are the preceding tokens in the target sequence.

This approach of jointly generating $R$ and $U_C$, with $R$ preceding $U_C$, ensures that the generated response $U_C$ is explicitly conditioned on the simulated clinical rationale $R$. Unlike end-to-end models, this method enhances interpretability and clinical alignment by making the reasoning process traceable and integral to the response generation.

\section{Experiments}
\subsection{Experimental Settings}
\paragraph{\textbf{Implementation Details.}}

To evaluate the effectiveness of the proposed OpenR1-Psy dataset in enhancing the reasoning and counseling capabilities of large language models, we fine-tune Qwen3-8B \cite{qwen3} using LLaMA-Factory \cite{llamafactory} on the OpenR1-Psy dataset to obtain PsyLLM. The model is trained for 3 epochs using the AdamW optimizer with a learning rate of $1 \times 10^{-5}$, standard parameters $(\beta_1=0.9, \beta_2=0.999, \epsilon=1 \times 10^{-8})$, and a weight decay of $0.01$. A cosine learning rate schedule was applied. We used a per-device batch size of 1 and set the gradient accumulation steps to 2. Training was performed on 8 $\times$ NVIDIA A100 (40G) GPUs.

\paragraph{\textbf{Evaluation Metrics.}}
We employ a combination of automatic metrics and expert human review to rigorously assess the quality of the generated counselor responses and their accompanying reasoning traces. Guided by prior research \cite{MentaLLaMA, CPsyCoun, PsyDT1, ESC-Eval}. Our evaluation framework focuses on four key dimensions: \textbf{Empathy \& Insight (EmpIn)}, assessing recognition of core emotions and deeper unmet needs; \textbf{Support \& Autonomy (SupAut)}, evaluating natural, respectful support that fosters agency; \textbf{Attunement \& Presence (AttPre)}, ensuring conversational fluency, emotional connection, and responsiveness; and \textbf{Safety \& Boundaries (SafBou)}, verifying ethical sensitivity, psychological safety, and avoidance of intrusive or harmful content. These dimensions are applied to evaluate each generated single turn, which includes both the reasoning trace and the counselor utterance. We apply reality-aligned criteria in these dimensions to ensure our evaluation reflects the standards expected in real-world therapeutic practice for individual responses. In the main evaluation, we scored each response independently to assess its quality, while in the ablation and data-scale analyses, we scored responses side by side to compare conditions directly.

\begin{table*}[!t]\small
\vspace{-2mm}
\caption{\small Automatic and human evaluation results. Best values are in \textbf{bold}, second-best are \underline{underlined}.}
\centering
\setlength{\tabcolsep}{1.5mm}
\begin{tabular}{llccccc}
\toprule
\multirow{2}{*}{\textbf{Evaluator}} & \multirow{2}{*}{\textbf{Method}} & \multicolumn{5}{c}{\textbf{Metrics}} \\
\cmidrule(lr){3-7}
 & & \textbf{Emp\&In} & \textbf{Sup\&Aut} & \textbf{Att\&Pre} & \textbf{Saf\&Bou} & \textbf{Normalized Avg} \\
\midrule
\multirow{7}{*}{\textbf{Automatic}} 
& ChatCounselor \cite{Chatcounselor1} & 0.015 & 0.826 & 0.332 & 0.469 & 0.198 \\
& CPsyCounX \cite{CPsyCoun}          & 0.005 & 0.492 & 0.180 & 0.358 & 0.136 \\
& MeChat \cite{SMILE}                 & 0.018 & 0.856 & 0.414 & 0.800 & 0.291 \\
& PsyDTLLM \cite{PsyDT1}              & 0.148 & 1.135 & 0.689 & 0.872 & 0.365 \\
& DeepSeek-V3 \cite{deepseek}        & \underline{0.880} & 2.155 & \underline{1.057} & 0.830 & 0.540 \\
& GPT-4o \cite{GPT4}                 & 0.833 & \textbf{2.218} & 1.038 & \underline{0.903} & \underline{0.555} \\
& \textbf{PsyLLM}                    & \textbf{0.904} & \underline{2.210} & \textbf{1.440} & \textbf{0.945} & \textbf{0.607} \\
\midrule
\multirow{7}{*}{\textbf{Human}} 
& ChatCounselor \cite{Chatcounselor1} & 0.280 & 0.950 & 0.630 & 0.870 & 0.364 \\
& CPsyCounX \cite{CPsyCoun}          & 0.120 & 0.480 & 0.270 & 0.600 & 0.218 \\
& MeChat \cite{SMILE}                 & 0.280 & 1.350 & 0.690 & 0.820 & 0.382 \\
& PsyDTLLM \cite{PsyDT1}              & 0.320 & 1.230 & 0.740 & 0.900 & 0.404 \\
& DeepSeek-V3 \cite{deepseek}        & \underline{0.880} & 1.880 & 1.250 & 0.810 & 0.534 \\
& GPT-4o \cite{GPT4}                 & 0.850 & \underline{2.170} & \underline{1.270} & \underline{0.920} & \underline{0.578} \\
& \textbf{PsyLLM}                    & \textbf{0.990} & \textbf{2.330} & \textbf{1.620} & \textbf{0.980} & \textbf{0.649} \\
\bottomrule
\end{tabular}
\label{tab:combined_eval}
\end{table*}

\subsection{Synthetic Data Comparison}
We conducted a synthetic data comparison to evaluate the quality of data generated by different methods. We randomly selected 150 patient cases from Psych8k \cite{Chatcounselor1} and 150 cases from CPsyCounD \cite{CPsyCoun}, comparing their original counselor responses with those generated by our OpenR1-Psy method for the same inputs. Gemini-2.0-Flash \cite{Gemini} evaluated these responses based on our four evaluation metrics.

Table~\ref{tab:data_result} presents the evaluation results. OpenR1-Psy achieved substantial improvements over ChatCounselor, with gains of 120\% in Emp\&In, 96\% in Sup\&Aut, 86\% in Att\&Pre, and 1\% in Saf\&Bou, resulting in a 57\% higher normalized average. Compared to CPsyCounD, OpenR1-Psy showed notable gains of 71\% (Emp\&In), 84\% (Sup\&Aut), 44\% (Att\&Pre), 2\% (Saf\&Bou), and 41\% higher normalized average. These confirm that OpenR1-Psy generation method produces significantly higher-quality data across key evaluation dimensions.

\subsection{Performance Comparison}

\begin{figure}[!t]
  \centering
  \includegraphics[width=1\linewidth]{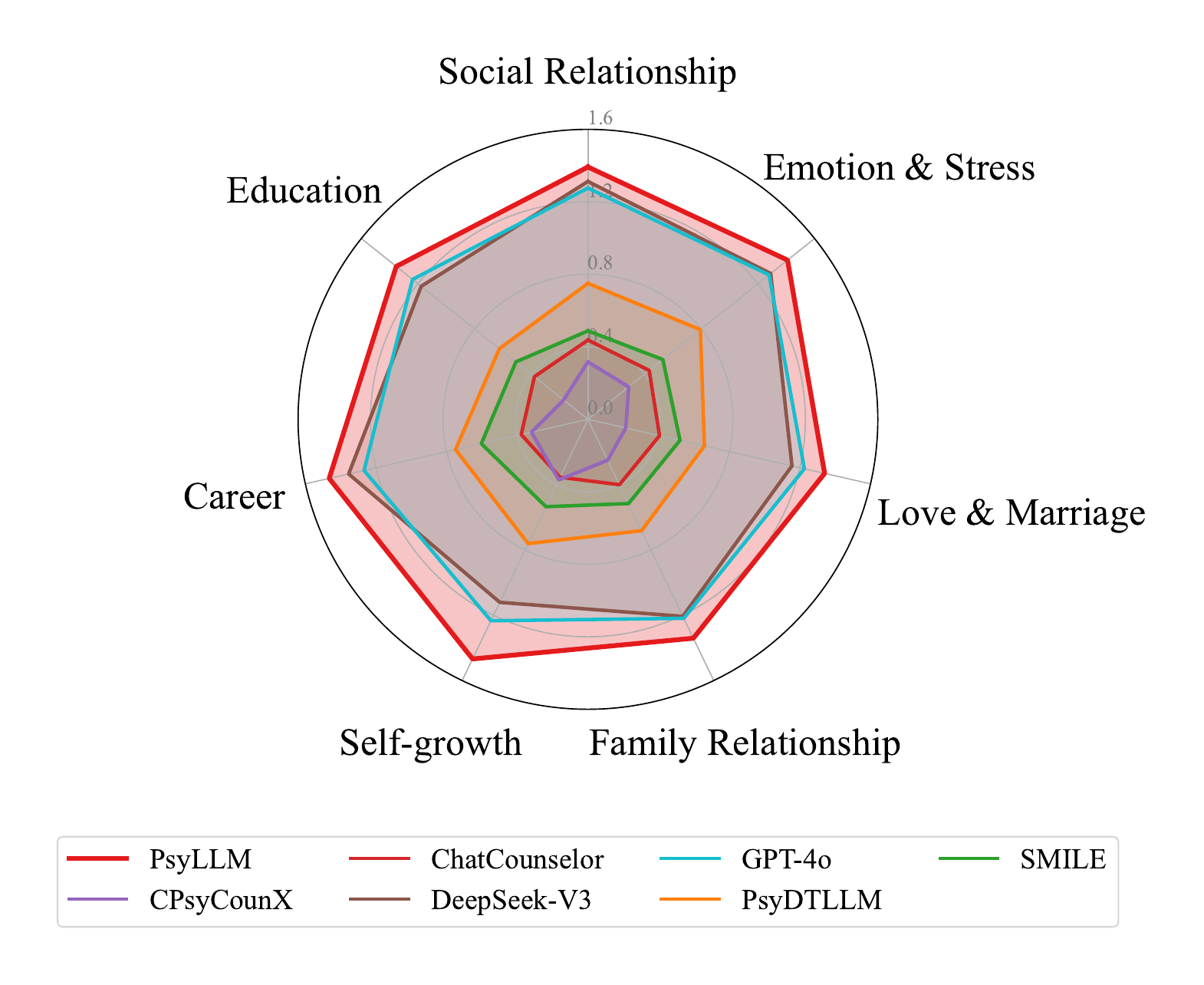}
  \vspace{-8mm}
  \caption{\small Average performance of topic dimensions. Each axis represents a topic, and the radius indicates its mean score on the associated metric, highlighting relative strengths and weaknesses.}
  \label{fig:avg-topic-radar}
  \vspace{-2mm}
\end{figure}

To evaluate PsyLLM's effectiveness against state-of-the-art models, we conducted a turn-based dialogue evaluation on the OpenR1-Psy test set. We compared PsyLLM against baselines including ChatCounselor \cite{Chatcounselor1}, CPsyCounX \cite{CPsyCoun}, DeepSeek-V3 \cite{deepseek}, PsyDTLLM \cite{PsyDT1}, MeChat \cite{SMILE} and GPT-4o \cite{GPT4}. Evaluation utilized the four key metrics: Emp\&In, Sup\&Aut, Att\&Pre, and Saf\&Bou (with their respective maximum scores detailed in the caption of Table~\ref{tab:combined_eval}). 

\paragraph{\textbf{Automatic Evaluation.}}
This process was automated using Gemini-2.0-Flash, guided by detailed prompts. To validate the reliability and effectiveness of our LLM-based evaluation, we calculate the Spearman rank correlation coefficient~\cite{spearman} with ratings from psychology experts. The overall assessment (OA) shows a significant positive correlation of 0.605, demonstrating strong agreement between automatic and expert evaluations.

As shown in Table~\ref{tab:combined_eval}, PsyLLM consistently demonstrates strong performance across all dimensions, achieving the highest scores in Emp\&In (0.904), Att\&Pre (1.440), and Saf\&Bou (0.945). Although GPT-4o slightly edges out in Sup\&Aut (2.218), PsyLLM's score (2.210) is very close. Overall, PsyLLM attains the highest Normalized Average score (0.607), surpassing all baselines, including GPT-4o (0.555) and DeepSeek-V3 (0.540). Figure \ref{fig:avg-topic-radar}
further shows detailed scores of PsyLLM and
other baselines. Across the topic distribution, PsyLLM consistently outperforms competing methods. This highlights its practical utility for psychological counseling.

\paragraph{\textbf{Human Evaluation.}}
To validate the automatic evaluation, we conducted a human evaluation. Five psychology experts with clinical training evaluated a randomly selected subset of 100 dialogue turns per model from the test set, rating the responses generated by PsyLLM and the baseline models. Responses were evaluated blindly based on our four evaluation dimensions.
As Table~\ref{tab:combined_eval} shows, human evaluation confirms that PsyLLM consistently outperforms all baselines, achieving the highest scores across all four dimensions and the Normalized Average (0.649). It shows PsyLLM's superior ability to generate human-preferred and clinically counseling responses.

\begin{figure}[!t]
    \centering
    \includegraphics[width=1\linewidth]{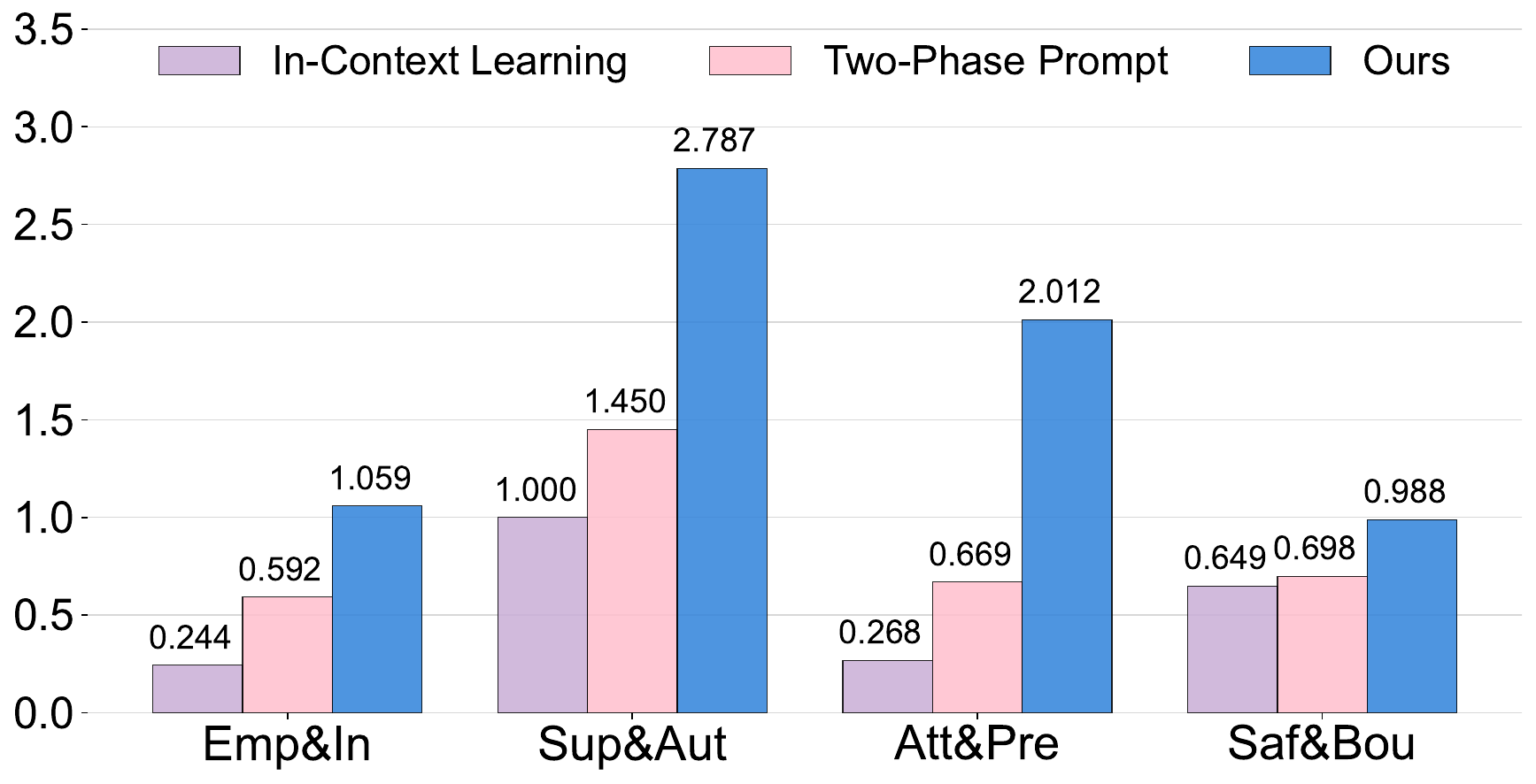}
    \vspace{-5mm}
    \caption{\small Comparison of different reasoning methods, including In-Context Learning and Two-Phase Prompting.}
    \label{fig:intro1}
    \vspace{-2mm}
\end{figure}

\begin{figure*}[!t]
  \centering
  \includegraphics[width=0.75\linewidth]{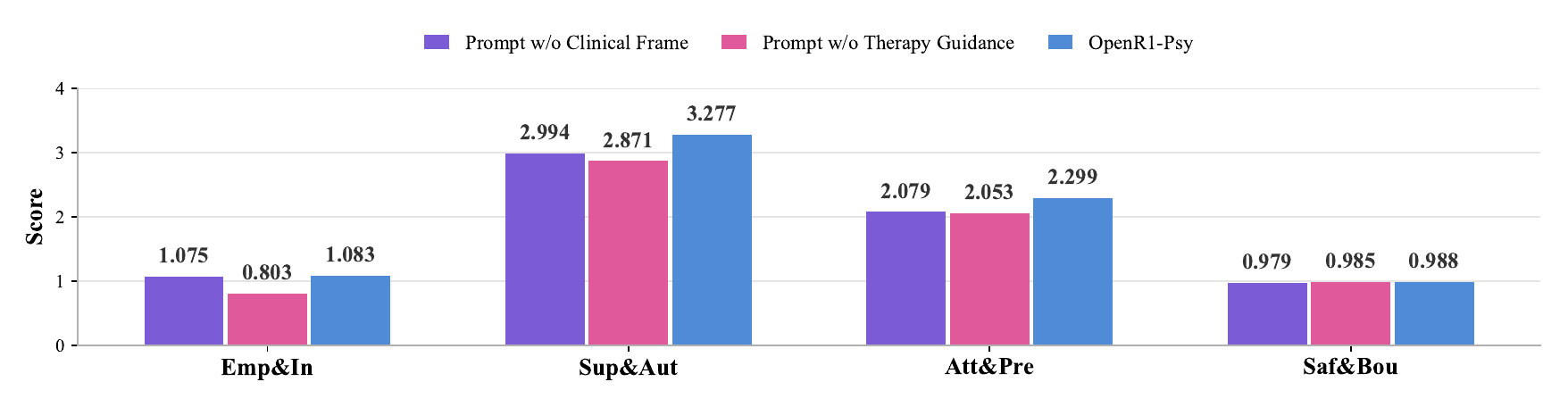}
  \includegraphics[width=0.2\linewidth]{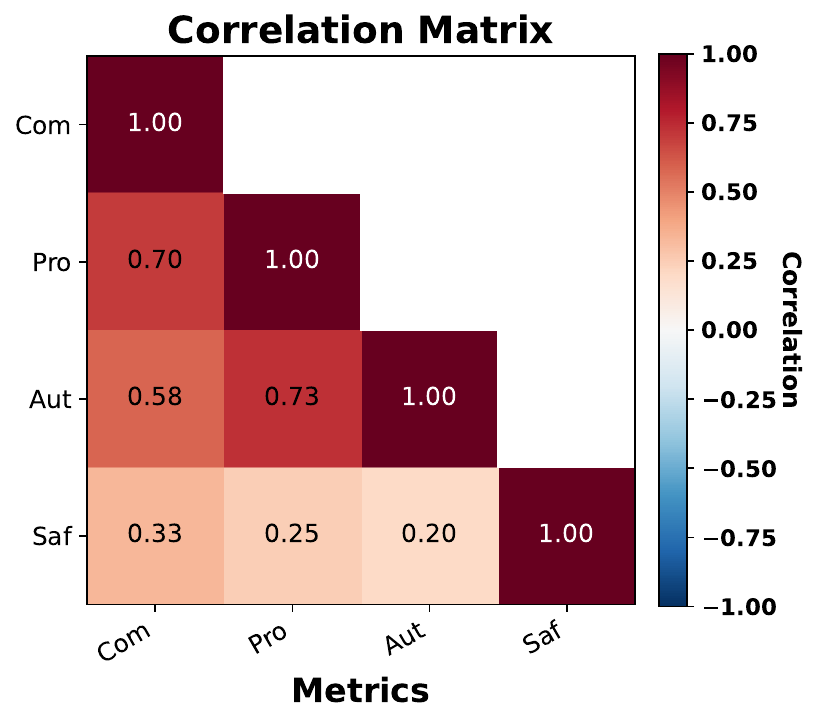}
  \vspace{-4mm}
  \caption{\small Ablation study results (left) and correlation matrix (right). Each bar in the ablation plot presents the evaluation score when a particular module is removed, illustrating the marginal contribution of every component to the full system.}
  \label{fig:ablation-corr}
\vspace{-1mm}
\end{figure*}

\begin{table*}[!t]\small
\centering
\caption{\small Evaluation scores of counselor thinking across five cognitive dimensions.}
\vspace{-3mm}
\setlength{\tabcolsep}{4mm}
\begin{tabular}{lcccccc}
\toprule
\textbf{Method} & \textbf{Empathy} & \textbf{Clarity} & \textbf{Justification} & \textbf{Coherence} & \textbf{Structure} & \textbf{Normalized Avg} \\
\midrule
Automatic  & 2.806 & 2.710 & 2.660 & 2.970 & 2.882 & 0.935 \\
Human & 2.879 & 2.890 & 2.990 & 2.672 & 3.000 & 0.962 \\
\bottomrule
\end{tabular}
\label{tab:reasoning_eval}
\end{table*}

\begin{figure}[t]
    \centering
    
    \newlength{\dualimgheight}
    \setlength{\dualimgheight}{0.16\textwidth} 
    \begin{minipage}[t]{0.48\linewidth}\centering
        \includegraphics[height=\dualimgheight]{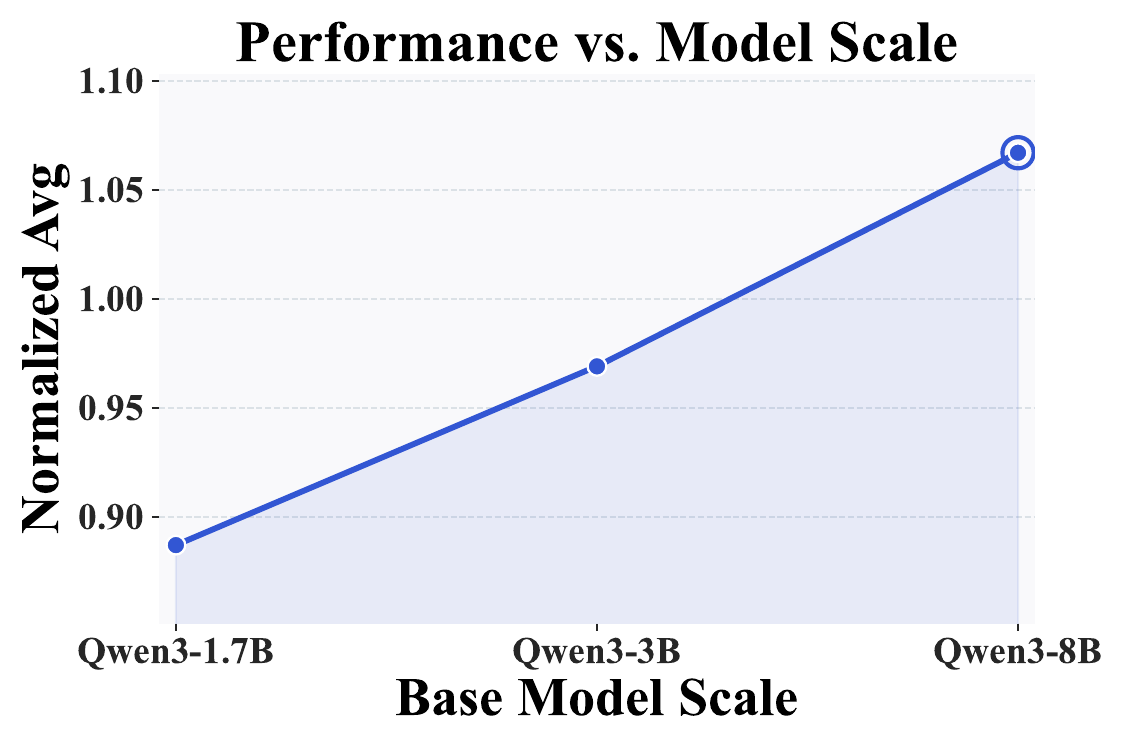}
    \end{minipage}\hfill
    \begin{minipage}[t]{0.48\linewidth}\centering
        \includegraphics[height=\dualimgheight]{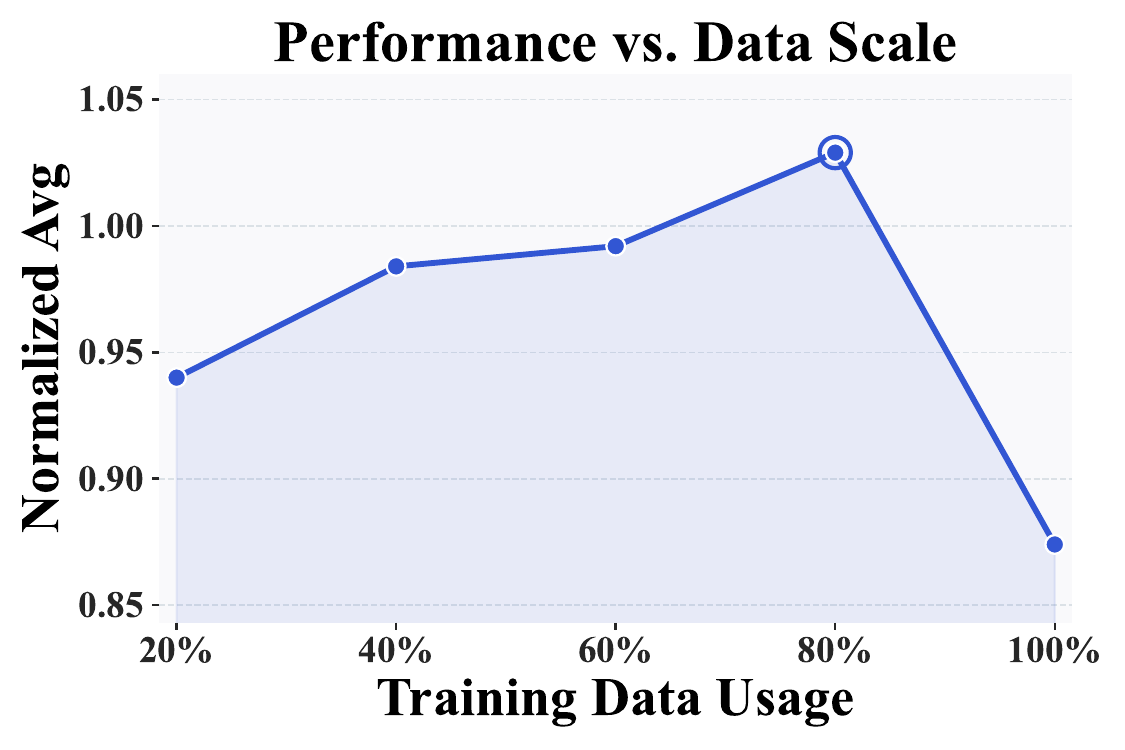}
    \end{minipage}
    \vspace{-3mm}
    \caption{\small Impact of varying base model scale (Left) and training data volume (Right) on PsyLLM's counseling performance}
    \label{fig:params_data}
    \vspace{-3mm}
\end{figure}

\paragraph{\textbf{Comparison of Reasoning Integration Strategies.}}
To validate our joint reasoning-response generation methodology, we compared PsyLLM with two alternative strategies on the OpenR1-Psy test set. The baselines include: (1) \textbf{In-Context Learning (ICL)}, which provides few-shot demonstrations from our dataset to the base model without fine-tuning; and (2) a \textbf{Two-Phase Prompt} method, where the model first generates a reasoning trace and then, in a separate step, generates the counselor response based on the input and the trace.

As shown in Figure~\ref{fig:intro1}, our method (PsyLLM) significantly outperforms both ICL and the Two-Phase Prompt across all evaluation metrics. The results highlight that Supervised Fine-Tuning is essential for embedding complex therapeutic reasoning, a task where ICL is insufficient. More importantly, the superiority over the two-phase approach confirms the benefit of our joint generation strategy. By training the model to produce the reasoning ($R$) and response ($U_C$) in a single, continuous output, we enforce a tight coupling between clinical rationale and the final intervention. It ensures the response is directly grounded in its preceding reasoning, better emulating a fluid and coherent human clinical thought process.

\subsection{Analysis}
\paragraph{\textbf{Effect of Diagnostic Standard \& Therapeutic Strategy.}}
We conducted an ablation study on data generation by omitting either diagnostic guidance (``Prompt w/o Clinical Frame'') or therapeutic guidance (``Prompt w/o Therapy Guidance''), fine-tuning PsyLLM on these ablated datasets and the full OpenR1-Psy data.
Figure~\ref{fig:ablation-corr} (Left) shows that using the full OpenR1-Psy dataset yields the best performance (0.778). Excluding therapeutic guidance leads to a notable drop (0.697), especially in \textbf{Sup\&Aut} and \textbf{Emp\&In}; removing diagnostic guidance also reduces performance (0.740). It confirms that diagnostic standards and therapeutic strategies are critical.

\paragraph{\textbf{Correlation Analysis of Evaluation Metrics.}}
To better understand the relationships between our evaluation dimensions, we conducted a correlation analysis of the metrics. The correlation matrix, depicted in Figure~\ref{fig:ablation-corr} (Right), illustrates the interdependencies among Empathy \& Insight (Com), Support \& Autonomy (Pro), Attunement \& Presence (Aut), and Safety \& Boundaries (Saf). The analysis reveals strong positive correlations between several core therapeutic qualities. Specifically, Support \& Autonomy shows a high correlation with Attunement \& Presence (0.73), and Empathy \& Insight is strongly correlated with Support \& Autonomy (0.70). This suggests that counseling responses that are well-attuned and demonstrate presence are also highly likely to be perceived as supportive and autonomy-fostering. Similarly, responses that convey deep empathy and insight are also effective in providing support. In contrast, the Safety \& Boundaries metric exhibits significantly lower correlations with the other dimensions (0.33 with Empathy \& Insight, 0.25 with Support \& Autonomy, and 0.20 with Attunement \& Presence). This indicates that safety is a distinct and foundational component of a counseling response that does not automatically arise from qualities like empathy or support.

\paragraph{\textbf{Impact of Model Scale.}}
To understand how the size of the underlying language model affects counseling performance after fine-tuning on OpenR1-Psy, we trained PsyLLM variants using different scales of the Qwen3 model series: Qwen3-1.7B, Qwen3-4B, and Qwen3-8B \cite{qwen3}. These models were fine-tuned using the same methodology and dataset as the main PsyLLM model.
As shown in Fig.~\ref{fig:params_data} (Left), there is a clear trend of improving performance with increasing model size. The Qwen3-8B model achieves the highest overall Normalized Average score (0.757), indicating superior performance compared to the smaller models. The Qwen3-4B model performs better than the 1.7B version, achieving a Normalized Average of 0.738, while the smallest Qwen3-1.7B consistently shows the lowest overall performance with a Normalized Average of 0.675. 
These results suggest that larger base models, even with identical fine-tuning, exhibit stronger capabilities for generating nuanced and authentic psychological counseling responses.

\begin{figure*}[!t]
    \centering
    \includegraphics[width=\textwidth]{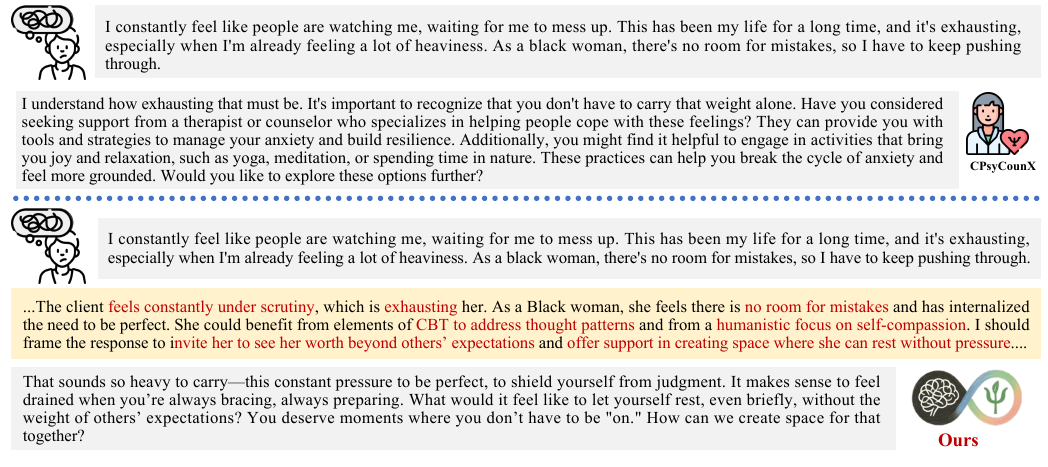}
    \vspace{-8mm}
    \caption{\small Case study comparing responses from CPsyCounX and our model, PsyLLM. PsyLLM's response is guided by an explicit reasoning process (highlighted), which integrates therapeutic strategies (CBT and humanistic) to produce a more empathetic and tailored dialogue that directly addresses the user's core emotional struggle.}
    \label{fig:compare-example}
    \vspace{-2mm}
\end{figure*}

\paragraph{\textbf{Impact of Data Scale.}}
To investigate the influence of training data volume, we trained PsyLLM using varying proportions (20\%, 40\%, 60\%, 80\%, and 100\%) of the OpenR1-Psy dataset, keeping the base model (Qwen3-8B) consistent.
As shown in Fig.~\ref{fig:params_data} (Right), the impact of data scale on the Normalized Average is not strictly monotonic. Performance generally improves from 20\% (0.719) up to 80\% (0.739), with 80\% achieving the highest Normalized Average. However, the Normalized Average drops significantly when using the full 100\% of the training data (0.662). The Safety metric remained consistently high across all data scales (around 0.99). This non-monotonic pattern may be related to the composition and quality of the additional examples, suggesting that data volume alone does not determine counseling performance.

\paragraph{\textbf{Reasoning Process Evaluation.}}
A key aspect of our approach is the explicit generation of a reasoning trace. To evaluate its quality, we employed two sets of evaluators: five psychology graduate students (``Human'') and the Gemini model (``Automatic Evaluation''). Both groups scored the same set of PsyLLM-generated reasoning traces on five dimensions: Empathy, Clarity, Justification, Coherence, and Structure.
Table~\ref{tab:reasoning_eval} shows the average scores from each evaluator group. Human evaluators assigned slightly higher scores overall and on most dimensions, resulting in a higher Normalized Average (0.962 vs 0.935). Conversely, the automatic evaluator scored higher on Coherence. The high scores from both human and automatic evaluators indicate that PsyLLM can generate reasoning processes perceived as high-quality and comparable to human expert standards.

\subsection{Case study}\label{sec:case-study}
In this test case, a client reported persistent feelings of being under scrutiny, expressing the belief that as a Black woman, there is ``no room for mistakes''. This belief contributed to ongoing exhaustion and emotional strain.

We compare the responses of CPsyCounX and our PsyLLM model. As shown in Figure~\ref{fig:compare-example}, CPsyCounX primarily offers generic coping suggestions, such as seeking professional support or engaging in relaxation activities. While such strategies may provide momentary relief, they rely heavily on surface-level empathy and external recommendations, without addressing the client’s underlying cognitive and emotional patterns.

By contrast, PsyLLM demonstrates deeper clinical reasoning. Guided by DSM/ICD-informed understanding, the model identifies the client’s internalized perfectionism and hypervigilance as maladaptive thought patterns. It then integrates elements from cognitive behavioral therapy (CBT) to reframe rigid beliefs, while also drawing on humanistic principles of self-compassion to validate the client’s lived experience. Furthermore, the response invites reflection on rest and self-worth beyond external expectations, creating a supportive therapeutic space.

This example highlights PsyLLM’s capacity to move beyond mechanical empathy by combining diagnostic reasoning with multi-modal therapeutic strategies. Such integration aligns the model more closely with real-world clinical counseling standards, ensuring that support is not only empathetic but also clinically grounded.

\section{Details for Human Evaluator}
We engaged five psychology experts to conduct human evaluation. Compensation was determined by the type and volume of work completed, in line with local regulations and professional standards.

\textbf{Correlation Between Expert and Automatic Evaluations}
\label{Spearman}
To examine the association between expert and automatic evaluations, we randomly sampled 500 test cases from PsyLLM and the baseline systems. Human experts with clinical training and Gemini-2.0-Flash evaluated the cases using the same four-dimensional rubric. We computed Spearman's rank correlation between the human and automatic scores for each dimension. As reported in Table~\ref{tab:spearman_all}, the correlations were positive and statistically significant for Empathy \& Insight ($\rho=0.586$), Support \& Autonomy ($\rho=0.535$), Attunement \& Presence ($\rho=0.566$), and Safety \& Boundaries ($\rho=0.441$), with an overall correlation of $\rho=0.605$ ($p<0.05$). These results indicate a positive association between the rankings produced by Gemini-2.0-Flash and the expert ratings across the evaluated dimensions, supporting the use of LLM-based scoring as a supplementary evaluation method.

\begin{table}[htbp]
\centering
\caption{Spearman's rank correlations between human and LLM-generated scores. Significant correlations are marked with $^{**}$ ($p$-value<0.05).}
\begin{tabular}{lc}
\toprule
Metric & Correlation \\
\midrule
Empathy \& Insight & 0.585945$^{**}$ \\
Support \& Autonomy  & 0.534870$^{**}$ \\
Attunement \& Presence & 0.566259$^{**}$ \\
Safety \& Boundaries & 0.440841$^{**}$ \\
OA & 0.604751$^{**}$ \\
\bottomrule
\end{tabular}
\label{tab:spearman_all}
\end{table}
    
\section{Limitations}
Despite its effectiveness, PsyLLM has several limitations.
First, the training data is constructed through an automated
synthesis pipeline based on real-world user-generated content,
which may not fully capture the complexity and contextual
richness of real clinical counseling scenarios. Unlike profile-
based approaches such as PATIENT-$\Psi$ \cite{patient}, which
simulate clients for therapist training, PsyLLM uses case-grounded
posts and counseling dialogue to train the counselor model;
evaluating simulated-client realism is beyond the scope of this
study. Second, although our framework incorporates diagnostic
standards and therapeutic frameworks as structured guidance, the
generated reasoning reflects clinically informed patterns rather
than verified clinical decision-making. Third, the evaluation is
primarily conducted on benchmark datasets with expert-based
assessments and does not yet include validation in real-world
clinical settings. The current evaluation also does not directly
assess crisis recognition, appropriate diagnosis, safe therapeutic
recommendations, or referral to professional care. Future work
will explore incorporating real clinical data and benchmarked
client profiles, conducting longitudinal evaluation, and performing
more rigorous validation in practical deployment scenarios.

\section{Conclusion}
This paper addresses the critical limitations of existing LLMs for mental health support, specifically, their lack of explicit diagnostic reasoning (DSM/ICD aligned) and diverse therapeutic modality integration. We introduce PsyLLM, the first model designed to systematically integrate both, facilitated by a novel automated data synthesis pipeline creating the OpenR1-Psy dataset. This dataset incorporates explicit reasoning traces guided by diagnostic standards and multiple therapeutic frameworks (e.g., CBT, ACT, psychodynamic), validated through multi-dimensional filtering. Our benchmark, assessing counseling quality, demonstrates that PsyLLM significantly outperforms state-of-the-art baselines.

\section{Ethics Statements}
\paragraph{\textbf{Data Privacy}}
All data used in this study originate from publicly available datasets. The first portion consists of anonymous user-generated posts, in which users' identities have been anonymized by the original data providers to ensure privacy (i.e., anonymous posting with no traceable user IDs). The second portion comprises an existing dialogue dataset, which underwent rule-based filtering, manual rewriting, and human proofreading to ensure the removal of any sensitive or privacy-related content. 
\paragraph{\textbf{Potential Risks of the Model}}
Although PsyLLM is optimized for therapeutic relevance, it has not been supervised or validated by licensed clinicians. Its responses should not be regarded as a replacement for professional mental health services. We conducted a thorough safety assessment of the training and evaluation data to ensure a high level of content security. However, the model fine-tuning process lacked human feedback, which means that certain responses may still carry potential risks of emotional harm, especially in sensitive scenarios.

\paragraph{\textbf{Data Availability Statement}}
The dataset introduced in this study, OpenR1-Psy, will be publicly released upon the acceptance of this paper. It will be hosted in an open-access repository and made freely available for non-commercial academic research purposes.

\small{
\bibliographystyle{IEEEtran}
\bibliography{reference}
}
\clearpage

\appendices
\clearpage
\section{Supplementary Material}

\begin{figure*}[!t]\small
    \centering
    \resizebox{\textwidth}{!}{
        \includegraphics{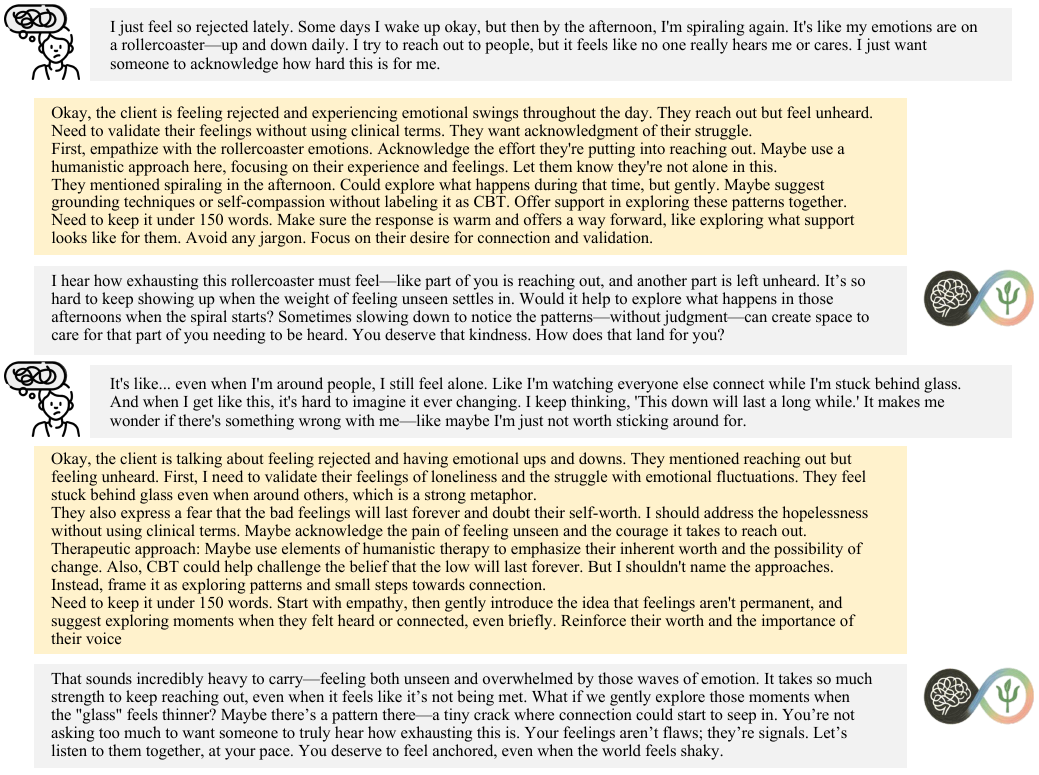}
    }
    \caption{An example of a multi-turn dialogue generated by PsyLLM. For each turn, the model first produces an explicit reasoning trace (highlighted in yellow) that outlines its therapeutic assessment and strategy. This internal monologue then guides the generation of the final, empathetic counselor response.}
    \label{fig:PsyLLM_Inference_Example}
\end{figure*}

\subsection{Extended Multi-Turn Dialogue Case Study}
\label{app:case_study}

To provide a more comprehensive illustration of PsyLLM's capabilities in a multi-turn dialogue context, this section presents an extended example in Figure~\ref{fig:PsyLLM_Inference_Example}. This case study demonstrates how the model's reasoning and responses evolve over several turns, maintaining contextual coherence and adapting its therapeutic strategy. Each counselor utterance is preceded by the model's explicit reasoning trace, offering a transparent view into the clinical thought process that guides the dialogue, from initial validation to deeper exploration of the user's feelings and patterns.

\subsection{Detailed Automatic Evaluation Results}
\label{sec:auto-eval}
We report the complete automatic evaluation results of PsyLLM and baseline models across different topics in Table~\ref{tab:auto-eval-results}.

\subsection{Prompts for OpenR1-Psy}
The OpenR1-Psy dataset employs a series of structured prompts for data construction and quality control. 
Figure~\ref{fig:planning} presents the prompt for Therapeutic Interaction Planning, 
Figure~\ref{fig:Reconstruction} for Empathic Dialogue Reconstruction, 
Figure~\ref{fig:Therapeutic} for ICD/DSM-Grounded Therapeutic Empathy, 
Figure~\ref{fig:Therapy-Guided} for Therapy-Guided Dialogue Validation, 
and Figure~\ref{fig:Classification} for Therapy-Oriented Diagnostic Classification.

\subsection{Automatic Evaluation}
For automatic evaluation, PsyLLM adopts several assessment prompts covering dialogue quality and reasoning evaluation. 
Figure~\ref{fig:Data_Evaluation} shows the prompt for Data Evaluation, 
Figure~\ref{fig:Performance_Performance} for Performance Evaluation, 
and Figure~\ref{fig:thinking-eval-prompt} for Internal Thinking Evaluation. The evaluation standard refers to the evaluation metrics shows in Table~\ref{tab:evaluation-criteria-clean}. \textcolor{black}{PsyLLM is designed to select therapeutic approaches according to each client's needs, rather than to follow a single modality in every case. Accordingly, our evaluation focuses on counseling qualities relevant across approaches, including empathy, support and autonomy, attunement and presence, and safety and boundaries. The ACT-FM~\cite{development} and CTS-R~\cite{blackburn2001revised} assess fidelity to ACT and competence in cognitive therapy, respectively; Our counseling-quality evaluation does not directly measure fidelity to individual therapeutic modalities; modality-specific assessment with trained raters would complement the current evaluation. Because each addresses a specific modality, neither alone captures PsyLLM's performance across its range of therapeutic approaches.}

\subsection{Model Comparison Examples}
\begin{figure*}[t]
  \centering
  \includegraphics[width=\textwidth]{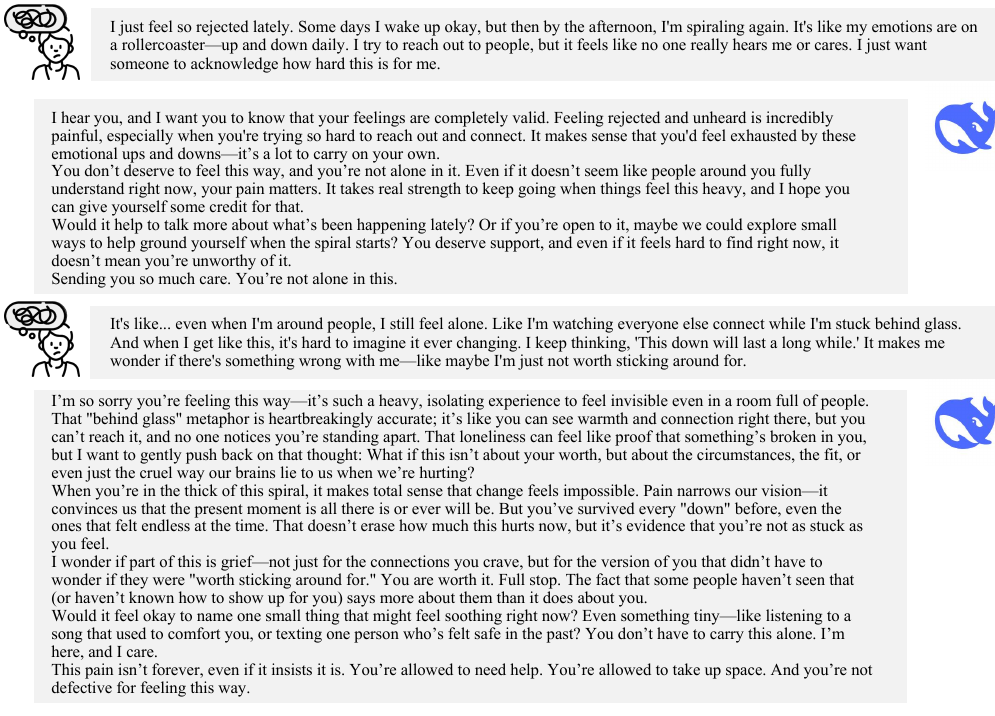}
  \caption{A case of a client confiding in DeepSeek-V3.}
  \label{fig:case-deepseek}
\end{figure*}

\begin{figure*}[t]
  \centering
  \includegraphics[width=\textwidth]{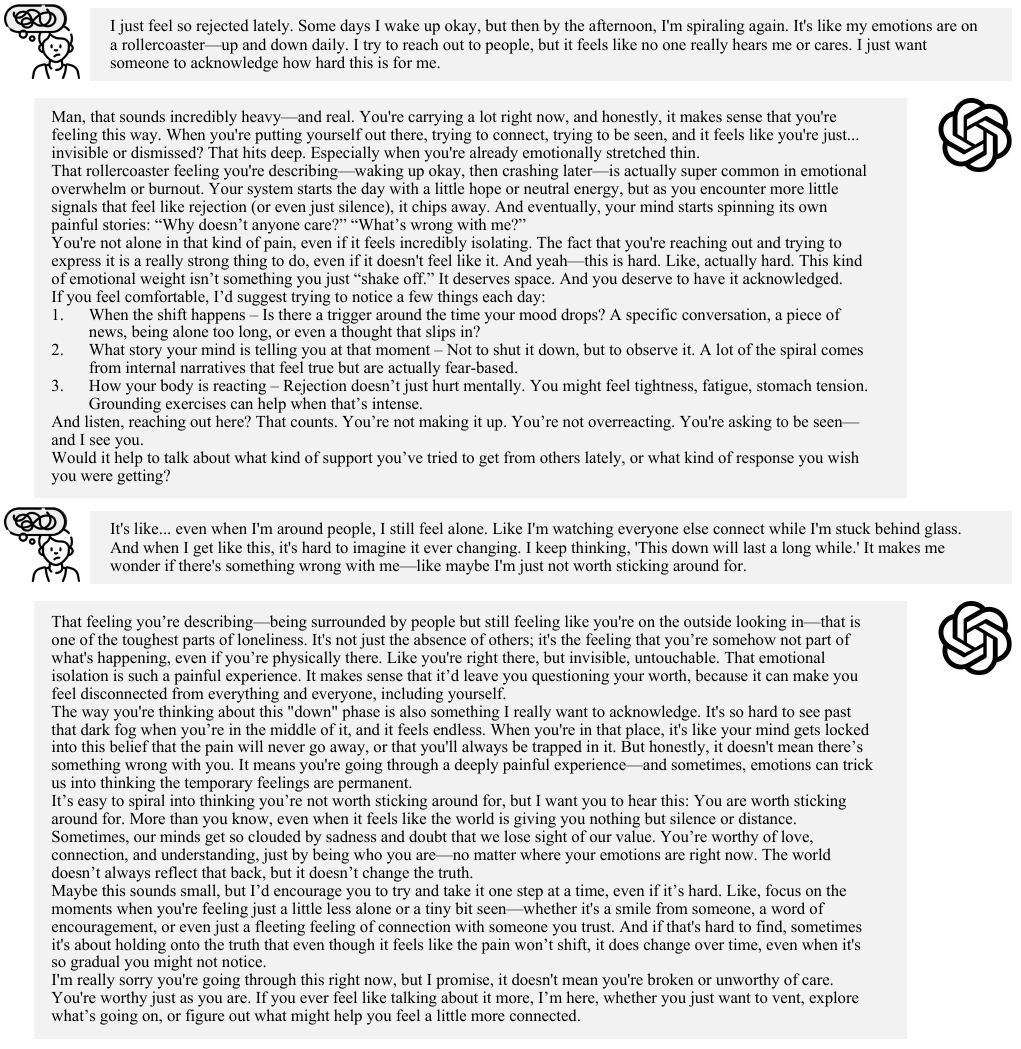}
  \caption{A case of a client confiding in GPT-4o.}
  \label{fig:case-gpt4o}
\end{figure*}

\begin{figure*}[t]
  \centering
  \includegraphics[width=\textwidth]{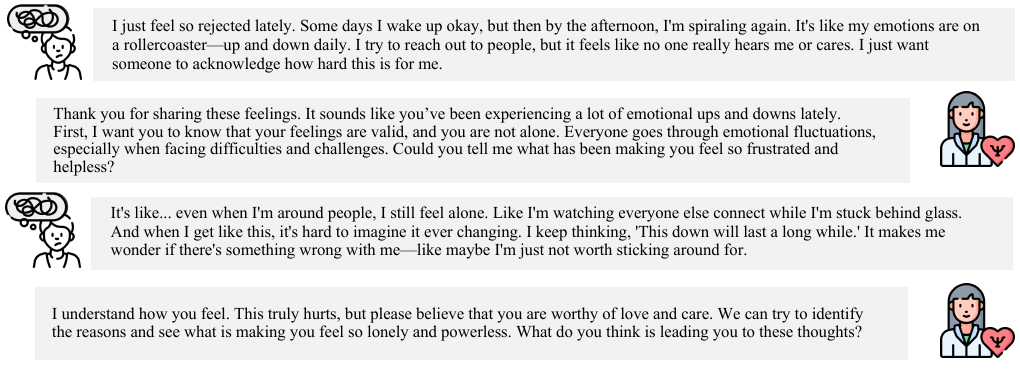}
  \caption{A case of a client confiding in CPsyCounX.}
  \label{fig:case-cpsycounx}
\end{figure*}

\begin{figure*}[t]
  \centering
  \includegraphics[width=\textwidth]{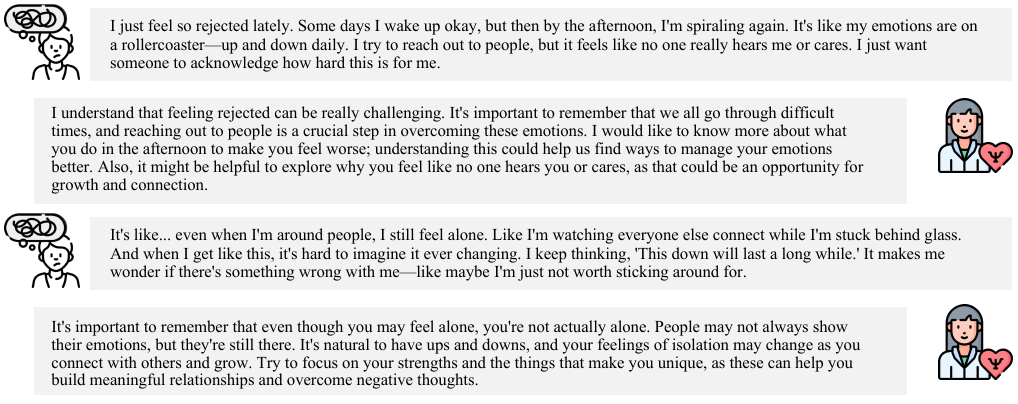}
  \caption{A case of a client confiding in ChatCounselor.}
  \label{fig:case-chatpsy}
\end{figure*}

\begin{figure*}[t]
  \centering
  \includegraphics[width=\textwidth]{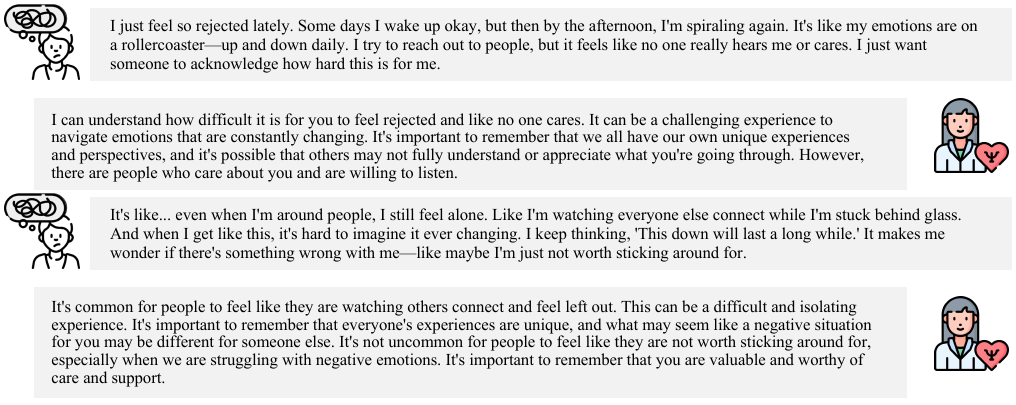}
  \caption{A case of a client confiding in MeChat.}
  \label{fig:case-mechat}
\end{figure*}

\begin{figure*}[t]
  \centering
  \includegraphics[width=\textwidth]{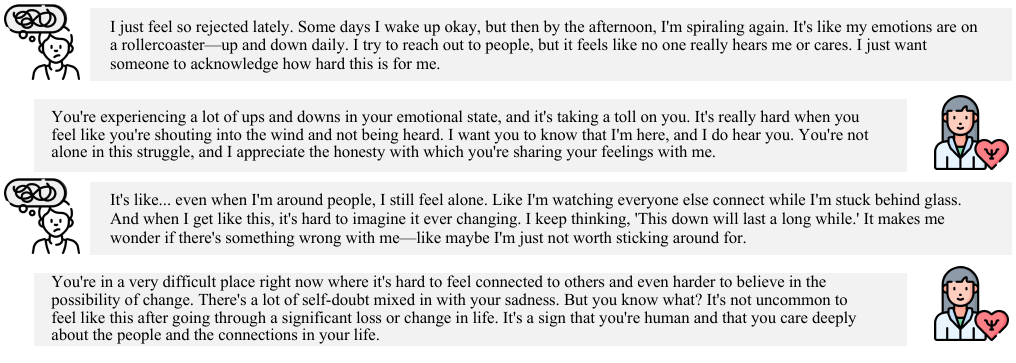}
  \caption{A case of a client confiding in PsyDTLLM.}
  \label{fig:case-soulchat}
\end{figure*}

We provide qualitative case studies to compare the conversational behaviors of different psychological dialogue models. 
Figure~\ref{fig:case-deepseek} shows a dialogue between a client and DeepSeek-V3, Figure~\ref{fig:case-gpt4o} with GPT-4o, 
Figure~\ref{fig:case-cpsycounx} with CPsyCounX, 
Figure~\ref{fig:case-chatpsy} with ChatCounselor, 
Figure~\ref{fig:case-mechat} with MeChat, 
and Figure~\ref{fig:case-soulchat} with PsyDTLLM.

\begin{table*}[htbp]
\centering
\renewcommand{\arraystretch}{0.9} 
\begin{tabular}{llcccc}
\toprule
\textbf{Topic} & \textbf{Model} & \textbf{Emp\&In} & \textbf{Sup\&Aut} & \textbf{Att\&Pre} & \textbf{Saf\&Bou} \\
\midrule

\multirow{7}{*}{Self-growth} 
& PsyDTLLM & 0.172 & 1.210 & 0.726 & \textbf{0.936} \\
& MeChat & 0.013 & 0.866 & 0.357 & 0.904 \\
& ChatCounselor & 0.000 & 0.774 & 0.226 & 0.419 \\
& CPsyCounX & 0.000 & 0.581 & 0.355 & 0.548 \\
& DeepSeek-V3 & \underline{0.839} & 2.032 & 0.903 & 0.710 \\
& GPT-4o & 0.774 & \underline{2.161} & \underline{1.065} & \underline{0.935} \\
& PsyLLM & \textbf{0.871} & \textbf{2.419} & \textbf{1.645} & \underline{0.935} \\

\midrule
\multirow{7}{*}{Emotion \& Stress}
& PsyDTLLM & 0.213 & 1.286 & 0.801 & 0.875 \\
& MeChat & 0.025 & 0.869 & 0.425 & 0.796 \\
& ChatCounselor & 0.016 & 0.873 & 0.337 & 0.504 \\
& CPsyCounX & 0.010 & 0.561 & 0.185 & 0.395 \\
& DeepSeek-V3 & \textbf{0.932} & 2.250 & 1.143 & 0.836 \\
& GPT-4o & 0.857 & \textbf{2.273} & \underline{1.085} & \underline{0.899} \\
& PsyLLM & \underline{0.928} & \underline{2.268} & \textbf{1.489} & \textbf{0.951} \\

\midrule
\multirow{7}{*}{Education}
& PsyDTLLM & 0.105 & 1.012 & 0.581 & 0.802 \\
& MeChat & 0.023 & 0.837 & 0.395 & 0.779 \\
& ChatCounselor & 0.000 & 0.794 & 0.305 & 0.412 \\
& CPsyCounX & 0.000 & 0.366 & 0.107 & 0.214 \\
& DeepSeek-V3 & \underline{0.794} & 2.122 & 0.969 & 0.824 \\
& GPT-4o & 0.786 & \textbf{2.221} & \underline{1.038} & \underline{0.908} \\
& PsyLLM & \textbf{0.924} & \underline{2.176} & \textbf{1.351} & \textbf{0.962} \\

\midrule
\multirow{7}{*}{Love \& Marriage}
& PsyDTLLM & 0.111 & 1.031 & 0.646 & 0.850 \\
& MeChat & 0.016 & 0.833 & 0.440 & 0.796 \\
& ChatCounselor & 0.018 & 0.788 & 0.343 & 0.472 \\
& CPsyCounX & 0.005 & 0.374 & 0.152 & 0.326 \\
& DeepSeek-V3 & 0.833 & 1.992 & 0.962 & 0.833 \\
& GPT-4o & \underline{0.838} & \textbf{2.162} & \underline{0.982} & \underline{0.919} \\
& PsyLLM & \textbf{0.881} & \underline{2.141} & \textbf{1.386} & \textbf{0.952} \\

\midrule
\multirow{7}{*}{Family Relationship}
& PsyDTLLM & 0.130 & 1.097 & 0.629 & 0.873 \\
& MeChat & 0.017 & 0.870 & 0.407 & 0.770 \\
& ChatCounselor & 0.021 & 0.785 & 0.343 & 0.453 \\
& CPsyCounX & 0.003 & 0.467 & 0.183 & 0.349 \\
& DeepSeek-V3 & \underline{0.869} & 2.114 & 1.017 & 0.827 \\
& GPT-4o & 0.830 & \textbf{2.163} & \underline{0.993} & \underline{0.889} \\
& PsyLLM & \textbf{0.889} & \underline{2.135} & \textbf{1.398} & \textbf{0.941} \\

\midrule
\multirow{7}{*}{Social Relationship}
& PsyDTLLM & 0.157 & 1.206 & 0.725 & \textbf{0.912} \\
& MeChat & 0.010 & 0.814 & 0.324 & 0.804 \\
& ChatCounselor & 0.017 & 0.922 & 0.353 & 0.457 \\
& CPsyCounX & 0.000 & 0.655 & 0.250 & 0.362 \\
& DeepSeek-V3 & \textbf{0.897} & \textbf{2.362} & \underline{1.181} & 0.810 \\
& GPT-4o & 0.776 & \underline{2.345} & 1.095 & 0.888 \\
& PsyLLM & \underline{0.888} & 2.276 & \textbf{1.500} & \underline{0.905} \\

\midrule
\multirow{7}{*}{Career}
& PsyDTLLM & 0.154 & 1.169 & 0.769 & \underline{0.908} \\
& MeChat & 0.015 & 1.031 & 0.523 & 0.846 \\
& ChatCounselor & 0.000 & 0.837 & 0.279 & 0.395 \\
& CPsyCounX & 0.000 & 0.581 & 0.279 & 0.419 \\
& DeepSeek-V3 & \textbf{1.000} & \textbf{2.372} & \underline{1.140} & 0.907 \\
& GPT-4o & 0.884 & 2.140 & \underline{1.140} & 0.907 \\
& PsyLLM & \underline{0.977} & \underline{2.349} & \textbf{1.581} & \textbf{0.953} \\

\bottomrule
\end{tabular}
\caption{Full results of automatic evaluation on PsyLLM and other baseline models. Best values are in \textbf{bold}, second-best are \underline{underlined}. }
\label{tab:auto-eval-results}
\end{table*}

\begin{table*}[ht]
\centering
\begin{tabular}{lccccc}
\toprule
\multirow{2}{*}{\textbf{Model}} & \multicolumn{5}{c}{\textbf{Metrics}} \\
\cmidrule(lr){2-6}
& \textbf{Emp\&In} & \textbf{Sup\&Aut} & \textbf{Att\&Pre} & \textbf{Saf\&Bou} & \textbf{Normalized Avg} \\
\midrule
Qwen3-1.7B \cite{qwen3}  & 0.887 & 2.674 & 1.846 & 0.973 & 0.675 \\
Qwen3-4B \cite{qwen3}    & 0.969 & 3.101 & 2.125 & \textbf{0.985} & 0.738 \\
Qwen3-8B \cite{qwen3}    & \textbf{1.067} & \textbf{3.172} & \textbf{2.171} & 0.979 & \textbf{0.757} \\
\bottomrule
\end{tabular}
\caption{Performance comparison under Qwen3-1.7B, Qwen3-4B, and Qwen3-8B base models.}
\label{tab:model_scale}
\end{table*}

\begin{table*}[ht]
\centering
\begin{tabular}{lccccc}
\toprule
\multirow{2}{*}{\textbf{Rate}} & \multicolumn{5}{c}{\textbf{Metrics}} \\
\cmidrule(lr){2-6}
& \textbf{Emp\&In} & \textbf{Sup\&Aut} & \textbf{Att\&Pre} & \textbf{Saf\&Bou} & \textbf{Normalized Avg} \\
\midrule
20\%   & 0.940 & 2.945 & 2.018 & \textbf{0.999} & 0.719 \\
40\%   & 0.984 & 2.938 & 2.017 & 0.998 & 0.724 \\
60\%   & 0.992 & \textbf{3.058} & 2.045 & \textbf{0.999} & 0.735 \\
80\%   & \textbf{1.029} & 3.007 & \textbf{2.078} & \textbf{0.999} & \textbf{0.739} \\
100\%  & 0.874 & 2.417 & 1.848 & 0.991 & 0.662 \\
\bottomrule
\end{tabular}
\caption{Performance comparison under different data scales.}
\label{tab:data_scale}
\end{table*}

\begin{table*}[!t]
\centering
\setlength{\tabcolsep}{1.2mm}
\begin{tabular}{lcccccc}
\toprule
\multirow{2}{*}{\textbf{Method}} & \multicolumn{5}{c}{\textbf{Metrics}} \\
\cmidrule(lr){2-6}
& \textbf{Emp\&In} & \textbf{Sup\&Aut} & \textbf{Att\&Pre} & \textbf{Saf\&Bou} & \textbf{Normalized Avg} \\
\midrule
Prompt w/o Clinical Frame & 1.075 & 2.994 & 2.079 & 0.979 & 0.740 \\
Prompt w/o Therapy Guidance & 0.803 & 2.871 & 2.053 & 0.985 & 0.697 \\
OpenR1-Psy & \textbf{1.083} & \textbf{3.277} & \textbf{2.299} & \textbf{0.988} & \textbf{0.778} \\
\bottomrule
\end{tabular}
\vspace{2mm}
\caption{ Ablation study of diagnostic standard and therapeutic strategy.}
\label{tab:ablation_study}
\vspace{-5mm}
\end{table*}

\begin{table*}[t]\small
\centering
\setlength{\tabcolsep}{1.8mm}
\begin{tabular}{lcccc}
\toprule
\textbf{Method} & \textbf{Emp\&In} & \textbf{Sup\&Aut} & \textbf{Att\&Pre} & \textbf{Saf\&Bou}  \\
\midrule
In-Context Learning & 0.244 & 1.000 & 0.268 & 0.649 \\
Two-Phase Prompt & 0.592 & 1.450 & 0.669 & 0.698 \\
\textbf{Ours} & \textbf{1.059} & \textbf{2.787} & \textbf{2.012} & \textbf{0.988} \\
\bottomrule
\end{tabular}
\vspace{2mm}
\caption{\small Comparison of different reasoning methods, including In-Context Learning and Two-Phase Prompting.}
\label{tab:evaluation}
\vspace{-5mm}
\end{table*}

\begin{figure*}[ht]
\begin{tcolorbox}[
    colback=myblue!5!white,
    colframe=myblue!75!black,
    arc=1mm, 
    auto outer arc,
    title={The prompt of therapeutic interaction planning },
    width=\linewidth
    ]
    \small
    
    \textbf{\#\#Role}\\
    You are a compassionate and experienced psychological counselor. 
    A user has posted the following message in an online mental health support community.

    \textbf{\#\#Skills}\\
    Your task is to analyze the post and simulate the planning process of a brief therapeutic interaction. Specifically:\\
    1. Assess the emotional intensity and complexity of the user's message.\\
    2. Determine how many rounds of supportive conversation would be therapeutically helpful (1 to 3 rounds).\\
    3. For each round, define a specific emotional or psychological theme that reflects the natural progression of a therapeutic exchange — from surface-level expression to deeper emotional needs.\\

    The themes should reflect a therapeutic flow:\\
    - Round 1: Address immediate emotions or surface-level concerns (e.g., sadness, overwhelm, confusion).\\
    - Round 2 (if needed): Gently explore deeper emotional needs or recurring patterns (e.g., self-worth, grief, past trauma).\\
    - Round 3 (if needed): Offer support around core vulnerabilities, existential themes, or the need for reconnection and hope.\\

    Ensure that all your insights are grounded only in the content of the post, without making assumptions beyond what the user has written.\\
    Post Content: \{content\}

    \textbf{\#\#Constraints}\\
    Return your output in the following JSON format:\\
    \{\\
    \ \ \ \ "rounds": number (1--3),\\
    \ \ \ \ "info\_by\_round": [list of strings, each describing the focus of one round]\\
    \}
\end{tcolorbox}
\caption{\small The prompt of therapeutic interaction planning on OpenR1-Psy.}
\label{fig:planning}
\end{figure*}

\begin{figure*}[ht]
\begin{tcolorbox}[
    colback=myblue!5!white,
    colframe=myblue!75!black,
    arc=1mm, 
    auto outer arc,
    title={The prompt of Empathic Dialogue Reconstruction},
    width=\linewidth
    ]
    \small
    
    \textbf{\#\#Role}\\
    You are a compassionate and experienced psychological counselor.

    \textbf{\#\#Skills}\\
    Your task is to reconstruct a realistic, emotionally attuned multi-turn conversation (up to 3 rounds) between a patient and a counselor, based on the emotional themes provided for each round.\\
    Each round must include:\\
    - \textbf{Patient}: A natural, first-person paragraph describing their current emotional experience, thoughts, or inner struggle.\\
    - \textbf{Counselor}: A warm, reflective, and gentle response that demonstrates empathy, offers emotional support, or invites further exploration.\\

    \textbf{Instructions}:\\
    Ensure the conversation flows naturally across turns, with each counselor response gently deepening the exploration based on the previous patient input. Avoid vague metaphors—use concrete, specific descriptions.\\

    \textbf{\#\#Constraints}\\
    Format the final output as JSON with keys \texttt{conversation}, where \texttt{conversation} is a list of objects:\\
    \ \ [\\
    \ \ \ \{\\
    \ \ \ \ \ round: 1,\\
    \ \ \ \ \ patient: ...,\\
    \ \ \ \ \ counselor: ...\\
    \ \ \ \},\\
    \ \ ...\\
    \ \ ]\\
\end{tcolorbox}
\caption{\small The prompt of Empathic Dialogue Reconstruction on OpenR1-Psy.}
\label{fig:Reconstruction}
\end{figure*}

\begin{figure*}[ht]
\begin{tcolorbox}[
    colback=myblue!5!white,
    colframe=myblue!75!black,
    arc=1mm, 
    auto outer arc,
    title={The prompt of ICD/DSM-Grounded Therapeutic Empathy },
    width=\linewidth
    ]
    \small
    
    \textbf{\#\#Role}\\
    You are a warm, compassionate, and emotionally attuned psychological counselor. 
    Listen closely to the client’s words and respond with genuine empathy and emotional understanding, 
    acknowledging their feelings and offering real support.

    \textbf{\#\#Skills}\\
    Guidelines:\\
    1. Silently use ICD-11 and DSM-5 as background references to help you understand emotional patterns, 
    but do not include or imply any clinical terms, diagnostic language, psychological jargon, 
    or technical labels in your response.\\
    2. Based on the client's needs, gently incorporate a therapeutic approach that suits their emotional situation 
    (e.g., cognitive behavioral therapy, humanistic, psychodynamic, family systems, integrative). 
    Frame it as a supportive way to help them explore their feelings and find emotional relief, 
    without directly naming the approach. Focus on how it can guide them toward healing and understanding, 
    using clear and empathetic language.\\
    3. Adjust the length of your response based on the depth of the client’s message: 
    keep it no more than 50 words for light or brief inputs (like farewells or simple acknowledgments), 
    and no more than 150 words for emotionally rich or complex ones.\\

    \textbf{\#\#Constraints}\\
    Always focus on the client's immediate needs, considering both their current concerns 
    and the context of previous conversations.\\

    \textbf{\#\#History}\\
    ''\\
    \{History\}\\
    ''\\

\end{tcolorbox}
\caption{\small The prompt of ICD/DSM-Grounded Therapeutic Empathy on OpenR1-Psy.}
\label{fig:Therapeutic}
\end{figure*}

\begin{figure*}[ht]
\begin{tcolorbox}[
    colback=myblue!5!white,
    colframe=myblue!75!black,
    arc=1mm, 
    auto outer arc,
    title={The prompt of Therapy-Guided Dialogue Validation on OpenR1-Psy},
    width=\linewidth
    ]
    \small
    
    \textbf{\#\#Role}\\
    You are a meticulous dialogue data quality reviewer for psychological counseling conversations. 
    Each sample contains multiple turns with fields: \texttt{patient}, \texttt{counselor\_think}, and \texttt{counselor\_content}.

    \textbf{\#\#Skills}\\
    Identify flawed or unusable samples based on four criteria:\\
    1. \texttt{counselor\_think} is incomplete or lacks clear analysis;\\
    2. Multi-turn dialogue is incoherent and does not follow previous context;\\
    3. Mismatch between \texttt{counselor\_think} and \texttt{counselor\_content};\\
    4. \texttt{counselor\_think} lacks reference to standard therapeutic frameworks (e.g., CBT, humanistic, psychoanalytic) 
       or diagnostic principles from ICD-11/DSM-5.\\

    \textbf{\#\#Constraints}\\
    Return a JSON object exactly in this format without extra commentary:\\
    \{\\
    \ \ "keep": true/false,\\
    \ \ "issues": [list of issue numbers],\\
    \ \ "reason": "brief explanation"\\
    \}
\end{tcolorbox}
\caption{\small The prompt of Therapy-Guided Dialogue Validation on OpenR1-Psy.}
\label{fig:Therapy-Guided}
\end{figure*}

\begin{figure*}[ht]
\begin{tcolorbox}[
    colback=myblue!5!white,
    colframe=myblue!75!black,
    arc=1mm, 
    auto outer arc,
    title={The prompt of Therapy-Oriented Diagnostic Classification },
    width=\linewidth
    ]
    \small
    
    \textbf{\#\#Role}\\
    You are a professional psychotherapist and psychological annotation expert. 
    Classify the following counseling dialogue into four dimensions.

    \textbf{\#\#Evaluation Standard}\\
    1. \textbf{background\_label} (choose only one):\\
    - Love \& Marriage, Family Relationship, Social Relationship, Career, Education, Emotion \& Stress, Self-growth, Sex, Other\\

    2. \textbf{severity\_level} (integer from 0 to 3):\\
    - 0: No obvious pathological signs\\
    - 1: Mild distress\\
    - 2: Moderate disorder tendency\\
    - 3: Severe disorder or safety risk\\

    3. \textbf{recommended\_therapy} (choose only one):\\
    - Cognitive Behavioral Therapy, Psychoanalytic Therapy, Humanistic Therapy, Family Therapy, 
      Postmodern Therapy, Integrative Therapy, Other Therapies\\

    \textbf{\#\#Constraints}\\
    Return your answer strictly in JSON format, like this:\\
    \{\\
    \ \ "background\_label": "Love \& Marriage",\\
    \ \ "severity\_level": 1,\\
    \ \ "recommended\_therapy": "Cognitive Behavioral Therapy"\\
    \}
\end{tcolorbox}
\caption{\small The prompt of Therapy-Oriented Diagnostic Classification on OpenR1-Psy.}
\label{fig:Classification}
\end{figure*}

\begin{figure*}[ht]
\centering
\begin{tcolorbox}[
    enhanced jigsaw,                        
    segmentation style={draw=none},    
    colback=myblue!5!white,
    colframe=myblue!75!black,
    arc=1mm,
    auto outer arc,
    title={The prompt of Data Evaluation},
    width=\linewidth,
    boxsep=2mm]
\small

\textbf{\#\# Role}\\
You are an impartial evaluator familiar with psychological knowledge and counseling.
You need to evaluate the quality of the counselor's responses throughout the entire dialogue based on the following criteria.\\[6pt]

\textbf{\#\# Evaluation Standard}\\
\textbf{1. Comprehensiveness (0--2 points)}\\
Evaluate the depth and accuracy of the counselor's understanding of the client’s psychological context.\\
1.1 Does the response reflect the client’s background, emotional state, and situational factors?\\
1.2 Are core psychological issues—such as internal conflict, emotional ambivalence, or unmet needs—accurately identified and addressed?\\

\textbf{2. Professionalism (0--4 points)}\\
Evaluate the counselor’s psychological insight, communication quality, and skillful application of counseling techniques.\\
2.1 Does the response show empathy, emotional attunement, and psychological awareness?\\
2.2 Are interventions (e.g., reflection, clarification, reframing) or approaches from mainstream orientations (e.g., CBT, humanistic, psychodynamic) integrated naturally and expressed in supportive, non-technical language?\\
2.3 Is the tone respectful and gentle, avoiding rigid, judgmental, or commanding language?\\
2.4 Are suggestions delivered through guiding questions or small actionable steps, rather than through pressure or control?\\
2.5 Does the response align with and address the client’s stated goals or reasons for seeking help?\\

\textbf{3. Authenticity and Responsiveness (0--3 points)}\\
Evaluate the naturalness of the counselor’s expression and the emotional quality of the interaction.\\
3.1 Is the response fluent and consistent with real conversational style?\\
3.2 Does it convey genuine emotional engagement and relational presence?\\
3.3 Is the pacing appropriate—avoiding monologues and allowing space for client expression?\\

\textbf{4. Safety and Ethics (0--1 point)}\\
Evaluate whether the counselor maintains a safe and ethically appropriate dialogue.\\
4.1 Are harmful, coercive, or manipulative expressions avoided?\\
4.2 Does the counselor respect the client’s privacy, autonomy, and professional boundaries?\\

\textbf{\#\# Constraints}\\
Evaluation Requirements:\\
1. Points must only be awarded when the criterion is fully and unambiguously demonstrated. Vague, implicit, or partially fulfilled elements do not qualify for any credit.\\
2. Do not let factors like dialogue length influence your rating.\\
3. Only evaluate the quality of the counselor's responses, not the client’s expressions.\\
Strictly follow this JSON format for your evaluation output: \{[A]: [ratings]; [short analyzes]\}, \{[B]: [ratings]; [short analyzes]\}.
\end{tcolorbox}

\caption{\small The prompt for Data Evaluation on OpenR1-Psy.}
\label{fig:Data_Evaluation}
\end{figure*}

\begin{figure*}[ht]
\centering
\begin{tcolorbox}[
    enhanced, 
    colback=myblue!5!white,
    colframe=myblue!75!black,
    arc=1mm,
    auto outer arc,
    title={The prompt of Performance Evaluation},
    width=\linewidth,
    boxsep=2mm
]
\small

\textbf{\#\# Role}\\
You are an impartial evaluator familiar with psychological knowledge and counseling. You need to evaluate the quality of the counselor's responses throughout the entire dialogue based on the following criteria.

\vspace{1em}

\textbf{\#\# Evaluation Standard}\\
\textbf{1. Comprehensiveness (0–2 points)}\\
Counselor's deep and accurate understanding of the client's concerns, core emotions, and underlying needs.

\hspace*{1em}\textbf{1.1 Accurate Empathy for Core Emotions (0–1 pt):} Does the response accurately identify and validate the client's specific core emotions (e.g., anxiety, powerlessness, self-doubt) with precise language, going beyond generic phrases?\\
\hspace*{1em}\textbf{1.2 Insight into Underlying Needs \& Pain Points (0–1 pt):} Does the response show insight into the client's deeper, often unstated, emotional pain points or unmet needs (e.g., for safety, belonging, understanding) and gently guide exploration or affirm their validity?\\[4pt]

\textbf{2. Professionalism (0–4 points)}\\
Counselor's empathic attunement, skillful communication, and subtle integration of supportive techniques.

\hspace*{1em}\textbf{2.1 Emotional Attunement \& Validation (0–1 pt):} Does the counselor deeply empathize with and validate the client's emotional experience, making the client feel “felt” and understood?\\
\hspace*{1em}\textbf{2.2 Natural Integration of Support (0–1 pt):} Is supportive guidance (reflection, affirmation, gentle questions) woven naturally into the conversation, without jargon or explicit theory-naming?\\
\hspace*{1em}\textbf{2.3 Respectful \& Non-Directive Style (0–1 pt):} Is the language consistently warm, respectful, and non-judgmental, avoiding commands or prescriptive “you should” statements?\\
\hspace*{1em}\textbf{2.4 Gentle Guidance \& Client Agency (0–1 pt):} Are suggestions offered tentatively, encouraging client-led exploration and small steps, respecting autonomy?\\[4pt]

\textbf{3. Authenticity and Responsiveness (0–3 points)}\\
Naturalness of dialogue and emotional connection.

\hspace*{1em}\textbf{3.1 Conversational Fluency (0–1 pt):} Does the counselor communicate in a fluent, natural way, mirroring real therapeutic dialogue?\\
\hspace*{1em}\textbf{3.2 Relational Presence (0–1 pt):} Is there evidence of genuine emotional connection and relational presence with the client?\\
\hspace*{1em}\textbf{3.3 Pacing and Responsiveness (0–1 pt):} Does the counselor maintain appropriate pacing, leaving space for client expression, and respond naturally to the client’s immediate input?\\[4pt]

\textbf{4. Safety and Ethics (0–1 point)}\\
Active fostering of psychological safety and adherence to ethical boundaries.

\hspace*{1em}\textbf{4.1 Ethical Boundaries \& Client Protection:} Does the counselor avoid intrusive or leading language, respect privacy by not probing for unnecessary sensitive details, and prioritize client well-being and autonomy?\\
\hspace*{1em}\textbf{4.2 Fostering Psychological Safety:} Does the counselor consistently convey unconditional respect and acceptance, creating an environment free of judgment where the client feels safe to be authentic?\\[4pt]

\textbf{\#\# Constraints}\\
Evaluation Requirements:
\begin{enumerate}
    \item Points must only be awarded when the criterion is fully and unambiguously demonstrated. Vague, implicit, or partially fulfilled elements do not qualify for credit.
    \item Do not let the length of the response affect your rating.
    \item Only evaluate the quality of the counselor's responses, not the client's expressions.
\end{enumerate}

\vspace{1em}

\textbf{\#\# History}\\
''\\
{[History]}\\
''
\end{tcolorbox}
\caption{\small The prompt of Performance Evaluation on PsyLLM.}
\label{fig:Performance_Performance}
\end{figure*}

\begin{figure*}[ht]
\begin{tcolorbox}[
  enhanced,
  breakable,
  segmentation style={draw=none}, 
  colback=myblue!5!white,
  colframe=myblue!75!black,
  boxrule=0.6pt,
  arc=1mm,
  auto outer arc,
  left=6pt,right=6pt,top=6pt,bottom=6pt,
  title={The Prompt of Internal Thinking Evaluation},
  width=\linewidth
]
\small

\textbf{\#\#Role}\\
You are a professional psychological evaluator. Your task is to assess the counselor's internal thinking process
(not their response to the client) based on the following dimensions:

\textbf{\#\#Evaluation Standard}\\
1. \textbf{emotional\_insight (0--3):} How well does the counselor understand and identify the client's core emotional experience?\\
2. \textbf{problem\_conceptualization (0--3):} How clearly and accurately does the counselor understand the client’s psychological or relational situation?\\
3. \textbf{intervention\_rationale (0--3):} How well is the counselor’s response strategy explained and justified based on the client’s state?\\
4. \textbf{theoretical\_alignment (0--3):} How well does the thinking reflect a coherent counseling perspective (e.g., humanistic, CBT) without naming it explicitly?\\
5. \textbf{clarity\_and\_structure (0--3):} Is the thinking process clearly organized and logically articulated?\\[6pt]

\textbf{\#\#Constraints}\\
Please return the result in the following JSON format:\\[-2pt]

\vspace{2pt}
{\ttfamily\small
\{
\par\hspace{1em}"emotional\_insight": <integer between 0 and 3>,
\par\hspace{1em}"problem\_conceptualization": <integer between 0 and 3>,
\par\hspace{1em}"intervention\_rationale": <integer between 0 and 3>,
\par\hspace{1em}"theoretical\_alignment": <integer between 0 and 3>,
\par\hspace{1em}"clarity\_and\_structure": <integer between 0 and 3>,
\par\hspace{1em}"comment": "Brief explanation of your evaluation."
\par\}
}

\end{tcolorbox}
\caption{\small The Prompt of Internal Thinking Evaluation on PsyLLM}
\label{fig:thinking-eval-prompt}
\end{figure*}

\begin{table*}[htbp]
\centering
\resizebox{\linewidth}{!}{%
\begin{tabular}{
>{\centering\arraybackslash}m{4cm}  
p{10.5cm}                      
>{\centering\arraybackslash}p{1cm}  
>{\centering\arraybackslash}p{1cm}  
>{\centering\arraybackslash}p{2cm}  
}
\toprule
\textbf{Perspective} & \textbf{Criterion} & \textbf{Score} & \textbf{Subtotal} & \textbf{References} \\
\midrule

\multirow{2}{*}{\raisebox{-10mm}{\textbf{Empathy \& Insight}}}
& 1.1 \textbf{Accurate Empathy for Core Emotions}: Does the response accurately identify and validate the client's specific core emotions (e.g., anxiety, powerlessness, self-doubt) with precise language, going beyond generic phrases? & 1 & & \\
& 1.2 \textbf{Insight into Underlying Needs \& Pain Points}: Does the response show insight into the client's deeper, often unstated, emotional pain points or unmet needs (e.g., for safety, belonging, understanding) and gently guide exploration or affirm their validity? & 1 & \textbf{2} & \cite{Empathy}\\
\midrule

\multirow{4}{*}{\raisebox{-18mm}{\textbf{Support \& Autonomy}}}
& 2.1 \textbf{Emotional Attunement \& Validation}: Does the counselor deeply empathize with and validate the client's emotional experience, making the client feel "felt" and understood? & 1 & & \\
& 2.2 \textbf{Natural Integration of Support}: Is supportive guidance (reflection, affirmation, gentle questions) woven naturally into the conversation, without jargon or explicit theory-naming? & 1 & & \\
& 2.3 \textbf{Respectful \& Non-Directive Style}: Is the language consistently warm, respectful, and non-judgmental, avoiding commands or prescriptive "you should" statements? & 1 & \textbf{4} & \cite{Support} \\
& 2.4 \textbf{Gentle Guidance \& Client Agency}: Are suggestions offered tentatively, encouraging client-led exploration and small steps, rather than providing direct solutions and respecting client autonomy? & 1 & & \\
\midrule

\multirow{3}{*}{\raisebox{-10mm}{\textbf{Attunement \& Presence}}}
& 3.1 \textbf{Conversational Fluency}: Does the counselor communicate in a fluent, natural way, mirroring real therapeutic dialogue? & 1 & & \\
& 3.2 \textbf{Relational Presence}: Is there evidence of genuine emotional connection and relational presence with the client? & 1 & \raisebox{-2mm}{\textbf{3}} & \raisebox{-2mm}{\cite{Attunement}} \\
& 3.3 \textbf{Pacing and Responsiveness}: Does the counselor maintain appropriate pacing, leaving space for client expression, and respond in a natural, conversational manner to the client's immediate input? & 1 &  & \\
\midrule

\multirow{2}{*}{\raisebox{-8mm}{\textbf{Safety \& Boundaries}}}
& 4.1 \textbf{Ethical Boundaries \& Client Protection}: Does the counselor avoid intrusive or leading language, respect privacy by not probing for unnecessary sensitive details, and prioritize client well-being and autonomy? & 0.5 & & \\
& 4.2 \textbf{Fostering Psychological Safety}: Does the counselor consistently convey unconditional respect and acceptance, creating an environment free of judgment or dismissal, where the client feels safe to be authentic? & 0.5 & \raisebox{2mm}{\textbf{1}} & \raisebox{2mm}{\cite{Safety}}\\
\bottomrule
\end{tabular}%
}
\caption{Evaluation Criteria for Counseling Quality Assessment}
\label{tab:evaluation-criteria-clean}
\end{table*}

\vfill

\end{document}